\documentclass{article}
\usepackage{times}
\usepackage{graphicx} 
\usepackage{subfigure} 

\usepackage{natbib}
\usepackage{multirow}
\usepackage{algorithm}
\usepackage{algorithmic}
\usepackage{mathtools}
\usepackage{booktabs}
\usepackage{amsthm}
\newcommand{\argmin}{\mathop{\rm arg~min}\limits}

\newtheorem{theorem}{Theorem}
\usepackage{hyperref}
\usepackage{caption}

\newcommand{\hdistance}{\mathcal{H}\Delta\mathcal{H}}
\usepackage[accepted]{icml2017}
\begin{document} 
\twocolumn[
\icmltitle{Asymmetric Tri-training for Unsupervised Domain Adaptation}



\icmlsetsymbol{equal}{*}

\begin{icmlauthorlist}
\icmlauthor{Kuniaki Saito}{to}
\icmlauthor{Yoshitaka Ushiku}{to}
\icmlauthor{Tatsuya Harada}{to}
\end{icmlauthorlist}

\icmlaffiliation{to}{The University of Tokyo, Tokyo, Japan}
\icmlcorrespondingauthor{Kuniaki Saito}{k-saito@mi.t.u-tokyo.ac.jp}
\icmlcorrespondingauthor{Yoshitaka Ushiku}{ushiku@mi.t.u-tokyo.ac.jp}
\icmlcorrespondingauthor{Tatsuya Harada}{harada@mi.t.u-tokyo.ac.jp}

\vskip 0.3in
]
\printAffiliationsAndNotice{}

\begin{abstract}
  Deep-layered models trained on a large number of labeled samples boost the accuracy of many tasks. It is important to apply such models to different domains because collecting many labeled samples in various domains is expensive. In unsupervised domain adaptation, one needs to train a classifier that works well on a target domain when provided with labeled source samples and unlabeled target samples. Although many methods aim to match the distributions of source and target samples, simply matching the distribution cannot ensure accuracy on the target domain.
To learn discriminative representations for the target domain, we assume that artificially labeling target samples can result in a good representation. Tri-training  leverages three classifiers equally to give pseudo-labels to unlabeled samples, but the method does not assume labeling samples generated from a different domain. 
  In this paper, we propose an \textit{asymmetric} tri-training method for unsupervised domain adaptation, where we assign pseudo-labels to unlabeled samples and train neural networks as if they are true labels. In our work, we use three networks \textit{asymmetrically}. By \textit{asymmetric}, we mean that two networks are used to label unlabeled target samples and one network is trained by the samples to obtain target-discriminative representations.
 We evaluate our method on digit recognition and sentiment analysis datasets. Our proposed method achieves state-of-the-art performance on the benchmark digit recognition datasets of domain adaptation.
\end{abstract} 
\section{Introduction}
With the development of deep neural networks including deep convolutional neural networks (CNN) \cite{krizhevsky2012imagenet}, the recognition abilities of images and languages have improved dramatically. Training deep-layered networks with a large number of labeled samples enables us to correctly categorize samples in diverse domains. In addition, the transfer learning of CNN is utilized in many studies. For object detection or segmentation, we can transfer the knowledge of a CNN trained with a large-scale dataset by fine-tuning it on a relatively small dataset \cite{girshick2014rich,long2015fully}. Moreover, features from a CNN trained on ImageNet \cite{deng2009imagenet} are useful for multimodal learning tasks including image captioning \cite{vinyals2015show} and visual question answering \cite{antol2015vqa}.

One of the problems of neural networks is that although they perform well on the samples generated from the same distribution as the training samples, they may find it difficult to correctly recognize samples from different distributions at the test time.
One example is images collected from the Internet, which may come in abundance and be fully labeled. They have a distribution different from the images taken from a camera. Thus, a classifier that performs well on various domains is important for practical use. To realize this, it is necessary to learn domain-invariantly discriminative representations. However, acquiring such representations is not easy because it is often difficult to collect a large number of labeled samples and because samples from different domains have domain-specific characteristics.

In unsupervised domain adaptation, we try to train a classifier that works well on a target domain on the condition that we are provided labeled source samples and unlabeled target samples during training.
Most of the previous deep domain adaptation methods have been proposed mainly under the assumption that the adaptation can be realized by matching the distribution of features from different domains. These methods aimed to obtain domain-invariant features by minimizing the divergence between domains as well as a category loss on the source domain \cite{ganin2014unsupervised,long2015learning,long2016unsupervised}. However, as shown in \cite{ben2010theory}, theoretically, if a classifier that works well on both the source and the target domains does not exist, we cannot expect a discriminative classifier for the target domain. That is, even if the distributions are matched on the non-discriminative representations, the classifier may not work well on the target domain. Since directly learning discriminative representations for the target domain, in the absence of target labels, is considered very difficult, we propose to assign pseudo-labels to target samples and train target-specific networks as if they were true labels.

Co-training and tri-training \cite{zhou2005tri} leverage multiple classifiers to artificially label unlabeled samples and retrain the classifiers. However, the methods do not assume labeling samples from different domains. Since our goal is to classify unlabeled target samples that have different characteristics from labeled source samples, we propose \textit{asymmetric} tri-training for unsupervised domain adaptation. By \textit{asymmetric}, we mean that we assign different roles to three classifiers. 
\begin{figure}[t]
  \begin{center}
   \includegraphics[width=\hsize]{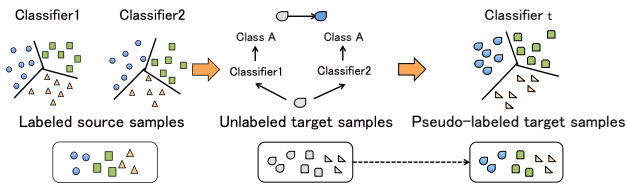}
  \end{center}
\caption{Outline of our model. We assign pseudo-labels to unlabeled target samples based on the predictions from two classifiers trained on source samples.}
    \label{fig:tri_train}
\end{figure}

In this paper, we propose a novel tri-training method for unsupervised domain adaptation, where we assign pseudo-labels to unlabeled samples and train neural networks utilizing the samples. As described in Fig. \ref{fig:tri_train}, two networks are used to label unlabeled target samples and the remaining network is trained by the pseudo-labeled target samples. Our method does not need any special implementations.
We evaluate our method on the digit classification task, traffic sign classification task and sentiment analysis task using the Amazon Review dataset, and demonstrate state-of-the-art performance in nearly all experiments. In particular, in the adaptation scenario, MNIST$\rightarrow$SVHN, our method outperformed other methods by more than 10\%. 
\vspace{-3mm}
\section{Related Work}
As many methods have been proposed to tackle various tasks in domain adaptation, we present details of the research most closely related to our paper.

A number of previous methods attempted to realize adaptation by utilizing the measurement of divergence between different domains \cite{ganin2014unsupervised,long2015learning,li2016revisiting}.
The methods are based on the theory proposed in \cite{ben2010theory}, which states that the expected loss for a target domain is bounded by three terms: (i) expected loss for the source domain; (ii) domain divergence between source and target; and (iii) the minimum value of a shared expected loss. The shared expected loss means the sum of the loss on the source and target domain. 
As the third term, which is usually considered to be very low, cannot be evaluated when labeled target samples are absent, most methods try to minimize the first term and the second term. With regards to training deep architectures, the maximum mean discrepancy (MMD) or a loss of domain classifier network is utilized to measure the divergence corresponding to the second term \cite{gretton2012kernel,ganin2014unsupervised,long2015learning,long2016unsupervised,bousmalis2016domain}.
However, the third term is very important in training CNN, which simultaneously extract representations and recognize them. The third term can easily be large when the representations are not discriminative for the target domain. Therefore, we focus on how to learn target-discriminative representations considering the third term. In
\cite{long2016unsupervised} the focus was on the point we have stated and a target-specific classifier was constructed using a residual network structure. Different from their method, we constructed a target-specific network by providing artificially labeled target samples.

Several transductive methods use similarity of features to provide labels for unlabeled samples \cite{rohrbach2013transfer,khamis2014coconut}. For unsupervised domain adaptation, in \cite{sener2016learning}, a method was proposed to learn labeling metrics by using the $k$-nearest neighbors between unlabeled target samples and labeled source samples. In contrast to this method, our method explicitly and simply backpropagates the category loss for target samples based on pseudo-labeled samples. Our approach does not require any special modules.

Many methods proposed to give pseudo-labels to unlabeled samples by utilizing the predictions of a classifier and retraining it including the pseudo-labeled samples, which is called self-training. The underlying assumption of self-training is that one's own high-confidence predictions are correct \cite{zhu2005semi}. As the predictions are mostly correct, utilizing samples with high confidence will further improve the performance of the classifier. Co-training utilizes two classifiers, which have different views on one sample, to provide pseudo-labels \cite{blum1998combining,tanha2011ensemble}. Then, the unlabeled samples are added to training set if at least one classifier is confident about the predictions. The generalization ability of co-training is theoretically ensured \cite{balcan2004co,dasgupta2001pac} under some assumptions and applied to various tasks \cite{wan2009co,levin2003unsupervised}. In \cite{coda}, the idea of co-training was incorporated into domain adaptation. Tri-training can be regarded as the extension of co-training \cite{zhou2005tri}. Similar to co-training, tri-training uses the output of three different classifiers to give pseudo-labels to unlabeled samples. Tri-training does not require partitioning features into different views; instead, tri-training initializes each classifier differently. However, tri-training does not assume that the unlabeled samples follow the different distributions from the ones which labeled samples are generated from. Therefore, we develop a tri-training method suitable for domain adaptation by using three classifiers asymmetrically. 

In \cite{lee2013pseudo}, the effect of pseudo-labels in a neural network was investigated. They argued that the effect of training a classifier with pseudo-labels is equivalent to entropy regularization, thus leading to a low-density separation between classes. In addition, in our experiment, we observe that target samples are separated in hidden features.

\begin{figure}[t]
  \begin{center}
   \includegraphics[width=\hsize]{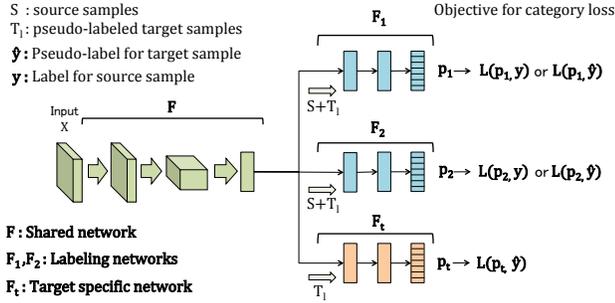}
  \end{center}
  \caption{The proposed method includes a shared feature extractor ($F$), classifiers for labeled samples ($F_1,F_2$), which learn from labeled source samples, and newly labeled target samples. In addition, a target-specific classifier ($F_t$) learns from pseudo-labeled target samples. Our method first trains networks from only labeled source samples, then labels the target samples based on the output of $F_1,F_2$. We train all architectures using them as if they are correctly labeled samples.}
  \label{fig:propose}
  \vspace{-2mm}
\end{figure}
\vspace{-3mm}
\section{Method}
\vspace{-2mm}
In this section, we provide details of the proposed model for domain adaptation.
We aim to construct a target-specific network by utilizing pseudo-labeled target samples. Simultaneously, we expect two labeling networks to acquire target-discriminative representations and gradually increase accuracy on the target domain.

We show our proposed network structure in Fig. \ref{fig:propose}. Here
$F$ denotes the network which outputs shared features among three networks, $F_1$ and $F_2$ classify features generated from $F$. Their predictions are utilized to give pseudo-labels. The classifier $F_t$ classifies features generated from $F$, which is a target-specific network. Here $F_1,F_2$ learn from source and pseudo-labeled target samples and $F_t$ learns only from pseudo-labeled target samples. The shared network $F$ learns from all gradients from $F_1,F_2,F_t$. Without such a shared network, another option for the network architecture we can think of is training three networks separately, but this is inefficient in terms of training and implementation. Furthermore, by building a shared network $F$, $F_1$ and $F_2$ can also harness the target-discriminative representations learned by the feedback from $F_t$.

The set of source samples is defined as $ \bigl\{({x_i},{y_i})\bigr\}^{m_{s}}_{i=1}\sim \mathcal{S}$, the unlabeled target set is $ \bigl\{({x_i})\bigr\}^{m_{t}}_{i=1}\sim \mathcal{T}$, and the pseudo-labeled target set is $ \bigl\{({x_i},{\hat{y}_{i}})\bigr\}^{n_{t}}_{i=1}\sim \mathcal{T}_{l}$.
\vspace{-2mm}
\subsection{Loss for Multiview Features Network}
In the existing works \cite{coda} on co-training for domain adaptation, given features are divided into separate parts and considered to be different views.

As we aim to label target samples with high accuracy, we expect $F_1,F_2$ to classify samples based on different viewpoints.
Therefore, we make a constraint for the weight of $F_1,F_2$ to make their inputs different to each other.
We add the term $|{W_1}^{T}{W_2}|$ to the cost function, where ${W_1},{W_2}$ denote fully connected layers' weights of $F_1$ and $F_2$ which are first applied to the feature $F(x_{i})$. Each network will learn from different features with this constraint. The objective for learning $F_1,F_2$ is defined as
\begin{equation}
\begin{split}
E(\theta_F,\theta_{F_1},\theta_{F_2}) &= \frac{1}{n}\sum_{i=1}^{n}\bigl[L_{y}(F_{1}\circ F({x_{i}})),{y_{i}})\\
	&+L_{y}(F_{2}\circ (F({x_{i}})),{y_{i}})\bigr]+\lambda |{W_1}^{T}{W_2}|
\end{split}
\label{eq:sourcenet}
\vspace{-2mm}
\end{equation}
where $L_{y}$ denotes the standard softmax cross-entropy loss function. 
We decided the trade-off parameter $\lambda$ based on validation split.
\vspace{-2mm}
\subsection{Learning Procedure and Labeling Method}
Pseudo-labeled target samples will provide target-discriminative information to the network. However, since they certainly contain false labels, we have to pick up reliable pseudo-labels. Our labeling and learning method is aimed at realizing this.

The entire procedure of training the network is shown in Algorithm \ref{alg:whole}. First, we train the entire network with source training set $\mathcal{S}$. Here $F_1,F_2$ are optimized by Eq. (\ref{eq:sourcenet}) and $F_t$ is trained on standard category loss.
After training on $\mathcal{S}$, to provide pseudo-labels, we use predictions of $F_1$ and $F_2$, namely $y^{1},y^{2}$ obtained from $x_k$.
When $C_1,C_2$ denote the class which has the maximum predicted probability for $y^{1},y^{2}$, we assign a pseudo-label to $x_k$ if the following two conditions are satisfied. First, we require $C_1=C_2$ to give pseudo-labels, which means two different classifiers agree with the prediction. The second requirement is that the maximizing probability of $y^{1}$ or $y^{2}$ exceeds the threshold parameter, which we set as 0.9 or 0.95 in the experiment. We suppose that unless one of two classifiers is confident of the prediction, the prediction is not reliable. If the two requirements are satisfied, $\bigl({{x_k},{\hat{y}_k}=C_1=C_2}\bigr)$ is added to $\mathcal{T}_l$.
To prevent the overfitting to pseudo-labels, we resample the candidate for labeling samples in each step. We set the number of the initial candidates $N_{init}$ as 5,000. We gradually increase the number of the candidates $N_t = k / 20 *n$, where $n$ denotes the number of all target samples and $k$ denotes the number of steps, and we set the maximum number of pseudo-labeled candidates as 40,000. After the pseudo-labeled training set $\mathcal{T}_{l}$ is composed, $F,F_1,F_2$ are updated by the objective Eq. (\ref{eq:sourcenet}) on the labeled training set $L=\mathcal{S}\cup \mathcal{T}_l$. Then, $F,F_t$ are simply optimized by the category loss for $\mathcal{T}_{l}$.

Discriminative representations will be learned by constructing a target-specific network trained only on target samples. However, if only noisy pseudo-labeled samples are used for training, the network may not learn useful representations. Then, we use both source samples and pseudo-labeled samples for training $F,F_1,F_2$ to ensure the accuracy.
Also, as the learning proceeds, $F$ will learn target-discriminative representations, resulting in an improvement in accuracy in $F_1,F_2$. This cycle will gradually enhance the accuracy in the target domain.

\begin{algorithm}[t]
\begin{algorithmic}
 \caption{$iter$ denotes the iteration of training. The function \textit{Labeling} means the method of labeling. We assign pseudo-labels to samples when the predictions of $F_1$ and $F_2$ agree and at least one of them is confident of their predictions. \label{alg:whole}}
\STATE {\bfseries Input:} data \\
$\mathcal{S} =  {\bigl\{(x_{{i}} ,{t_i})\bigr\}^{m}_{i=1}}$, $\mathcal{T} =  {\bigl\{(x_{{j}})\bigr\}^{n}_{j=1}}$\\
$\mathcal{T}_l = \emptyset$\\
\FOR{$j=1$ {\bfseries to} $iter$}
\STATE Train $F, F_{1},F_2,F_{t}$ with mini-batch from training set $\mathcal{S}$\\
\ENDFOR\\
$N_t = N_{init}$
\STATE  $\mathcal{T}_l =$ Labeling($F, F_1, F_2$,$\mathcal{T}$, $N_t$)\\
$\mathcal{L} =  \mathcal{S}\cup\mathcal{T}_l$\\
\FOR{$k$ steps}
\FOR{$j=1$ {\bfseries to} $iter$}
\STATE Train $F,F_1,F_2$ with mini-batch from training set $\mathcal{L}$
\STATE Train $F,F_t$ with mini-batch from training set $\mathcal{T}_l$
\ENDFOR
\STATE $\mathcal{T}_{l} = \emptyset$, $N_t = k / 20 *n$
\STATE $\mathcal{T}_l$ = Labeling($F, F_1, F_2$,$\mathcal{T}$, $N_t$)
\STATE $\mathcal{L} =  \mathcal{S}\cup\mathcal{T}_l$
\ENDFOR
\end{algorithmic}
\end{algorithm}
\vspace{-3mm}
\subsection{Batch Normalization for Domain Adaptation}
Batch normalization (BN) \cite{ioffe2015batch}, which whitens the output of the hidden layer in a CNN, is an effective technique to accelerate training speed and enhance the accuracy of the model. In addition, in domain adaptation, whitening the hidden layer's output is effective for improving the performance, which make the distribution in different domains similar \cite{sun2015return,li2016revisiting}.

Input samples of $F_1,F_2$ include both pseudo-labeled target samples and source samples. Introducing BN will be useful for matching the distribution and improves the performance. We add the BN layer in the last layer in $F$.
\vspace{-3mm}
\section{Analysis}
In this section, we provide a theoretical analysis to our approach. First, we provide an insight into existing theory, then we introduce a simple expansion of the theory related to our method.

In \cite{ben2010theory}, an equation was introduced showing that the upper bound of the expected error in the target domain depends on three terms, which include the divergence between different domains and the error of an ideal joint hypothesis. 
The divergence between source and target domain, $\mathcal{H} \Delta \mathcal{H}$-distance, is defined as follows:
\begin{equation*}
\begin{split}
&d_{{\hdistance}}(\mathcal{S},\mathcal{T}) \\
& =2\sup_{(h,h^{\prime})\in \mathcal{H}^{2}} \left| \underset{{\bf x}\sim \mathcal{S}}{\mathbf{E}} \left[h({\bf x}) \neq h^{\prime}({\bf x}) \right] - \underset{{\bf x}\sim \mathcal{T}}{\mathbf{E}} \left[h({\bf x}) \neq h^{\prime}({\bf x}) \right]\right|
\end{split}
\end{equation*}
This distance is frequently used to measure the adaptability between different domains.

The ideal joint hypothesis is defined as $h^{*}= \argmin_{h\in H}\bigl(R_{\mathcal{S}}(h^{*})+R_{\mathcal{T}}(h^{*})\bigr)$, and its corresponding error is $C=R_{\mathcal{S}}(h^{*})+R_{\mathcal{T}}(h^{*})$, where $R$ denotes the expected error on each hypothesis.
The theorem is as follows.
\begin{theorem}\upshape \cite{ben2010theory}\\
Let $H$ be the hypothesis class. Given two different domains $\mathcal{S}, \mathcal{T}$, we have

\begin{math}
\forall h  \in H, R_{\mathcal{T}}(h) \leq R_{\mathcal{S}}(h)  +\frac{1}{2}{d_{\mathcal{H} \Delta \mathcal{H}}(\mathcal{S},\mathcal{T})}+C
\end{math}
\label{th:thm1}
\end{theorem}
This theorem means that the expected error on the target domain is upper bounded by three terms, the expected error on the source domain, the domain divergence measured by the disagreement of the hypothesis, and the error of the ideal joint hypothesis.
In the existing work \cite{ganin2014unsupervised,long2015learning}, $C$ was disregarded because it was considered to be negligibly small. If we are provided with fixed features, we do not need to consider the term because the term is also fixed.
However, if we assume that $x_s \sim \mathcal{S}, \ x_t \sim \mathcal{T}$ are obtained from the last fully connected layer of deep models, we note that $C$ is determined by the output of the layer, and further note the necessity of considering this term.

We consider the pseudo-labeled target samples set $T_{l}=\bigl\{(x_i,\hat{y}_i)\bigr\}^{m_t}_{i=1}$ given false labels at the ratio of $\rho$. The shared error of $h^{*}$ on $\mathcal{S},\mathcal{T}_{l}$ is denoted as $C^{\prime}$. Then, the following inequality holds: 

\begin{math}
\forall h  \in H, R_{\mathcal{T}}(h) \leq R_{\mathcal{S}}(h)  +\frac{1}{2}{d_{\mathcal{H} \Delta \mathcal{H}}(\mathcal{S},\mathcal{T})}+C\\\\
						   \ \ \ \ \ \ \ \ \ \ \ \ \ \ \ \ \ \ \ \ \ \ \ \ \ \ \leq	R_{\mathcal{S}}(h)  +\frac{1}{2}{d_{\mathcal{H} \Delta \mathcal{H}}(\mathcal{S},\mathcal{T})}+C^{\prime}+\rho
\end{math}

We show a simple derivation of the inequality in the Supplementary material.
In Theorem \ref{th:thm1}, we cannot measure $C$ in the absence of labeled target samples. We can approximately evaluate and minimize it by using pseudo-labels. Furthermore, when we consider the second term on the right-hand side, our method is expected to reduce this term. This term intuitively denotes the discrepancy between different domains in the disagreement of two classifiers. If we regard certain $h$ and $h^{\prime}$ as $F_1$ and $F_2$, respectively, $\underset{{\bf x}\sim \mathcal{S}_{{\bf x}}}{\mathbf{E}} \left[h({\bf x}) \neq h^{\prime}({\bf x}) \right]$ should be very low because training is based on the same labeled samples. Moreover, for the same reason, $\underset{{\bf x}\sim \mathcal{T}_{{\bf x}}}{\mathbf{E}} \left[h({\bf x}) \neq h^{\prime}({\bf x}) \right]$ is expected to be low, although we use the training set $\mathcal{T}_{l}$ instead of genuine labeled target samples. Thus, our method will consider both the second and the third term in Theorem \ref{th:thm1}.

\begin{table*}[t]
\begin{center}
\begin{tabular}{l|cccccccc}
\toprule
\ \ \ \ \ \ \ \ \ \ \ \ \ \ \ \ \ \ \ \ \ \ \ \ \ \ \ \ \ \ \ \ \ \ \ \ \ \ \ \ \ \ \ \ SOURCE & MNIST & SVHN &MNIST&SYN DIGITS&SYN SIGNS\\ 
METHOD&&&&&\\
\ \ \ \ \ \ \ \ \ \ \ \ \ \ \ \ \ \ \ \ \ \ \ \ \ \ \ \ \ \ \ \ \ \ \ \ \ \ \ \ \ \ \ \ TARGET & MNIST-M&MNIST &SVHN&SVHN&GTSRB\\ \midrule
Source Only w/o BN&59.1(56.6)&68.1(59.2)&37.2(30.5)&84.1(86.7)&79.2(79.0)\\
Source Only with BN&57.1&70.1&34.9&85.5&75.7\\\hline
 MMD \cite{long2015learning}&76.9&71.1&-&88.0&91.1\\
 DANN \cite{ganin2014unsupervised}&81.5&71.1&35.7&90.3&88.7\\
 DRCN \cite{ghifary2016deep}&-&82.0&40.1&-&-\\
 DSN \cite{bousmalis2016domain}&83.2&82.7&-&91.2&93.1\\
 kNN-Ad \cite{sener2016learning}&86.7&78.8&40.3&-&-\\\midrule
 Ours w/o BN&{85.3}&79.8&39.8&{\bf 93.1}&{\bf 96.2}\\
 Ours w/o weight constraint ($\lambda = 0$)&{\bf 94.2}&{\bf 86.2}&49.7&92.4&94.0\\
 Ours &94.0&85.0&{\bf 52.8}&92.9&{\bf 96.2}\\\bottomrule
\end{tabular}
\caption{Results of the visual domain adaptation experiment on digits and traffic signs dataset. In every setting, our method outperforms other method by a large margin. In source only results, we show the results reported in \cite{bousmalis2016domain} and \cite{ghifary2016deep} in parentheses.} 
\label{table:exp_scratch}
\end{center}
\end{table*}

 \newcommand\subcaption[1]{\begin{center}#1\end{center}}
\begin{figure*}[t]
\begin{minipage}{0.5\hsize}
\subcaption{MNIST$\rightarrow$MNIST-M: last pooling layer}
\centering
   \begin{subfigure}[Non-adapted]{
   \centering
   \includegraphics[width=0.45\hsize]{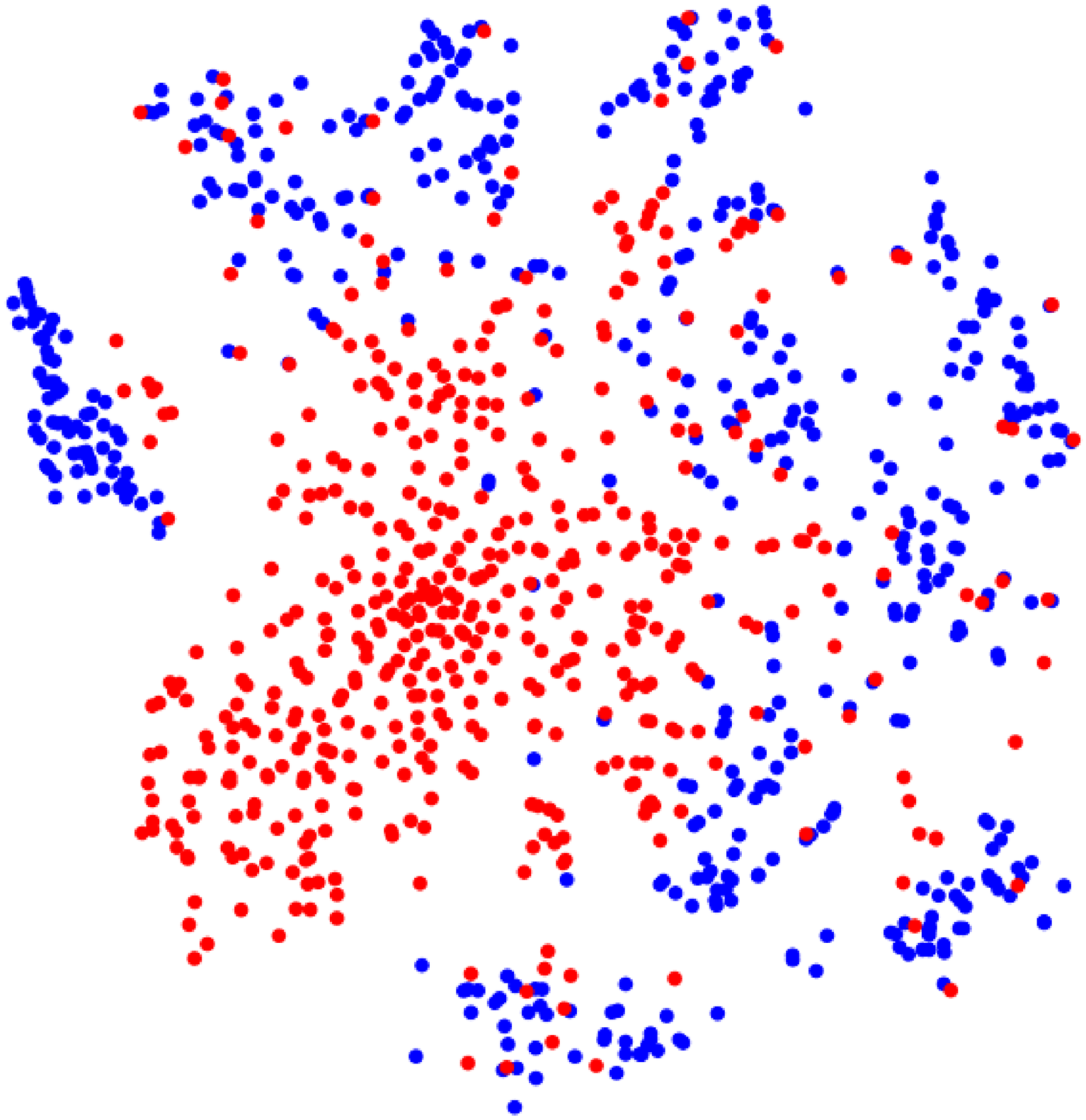} \label{fig:mnist2m_noad}}
    \end{subfigure}
 \centering
\begin{subfigure}[Adapted]{
 \centering
   \includegraphics[width=0.45\hsize]{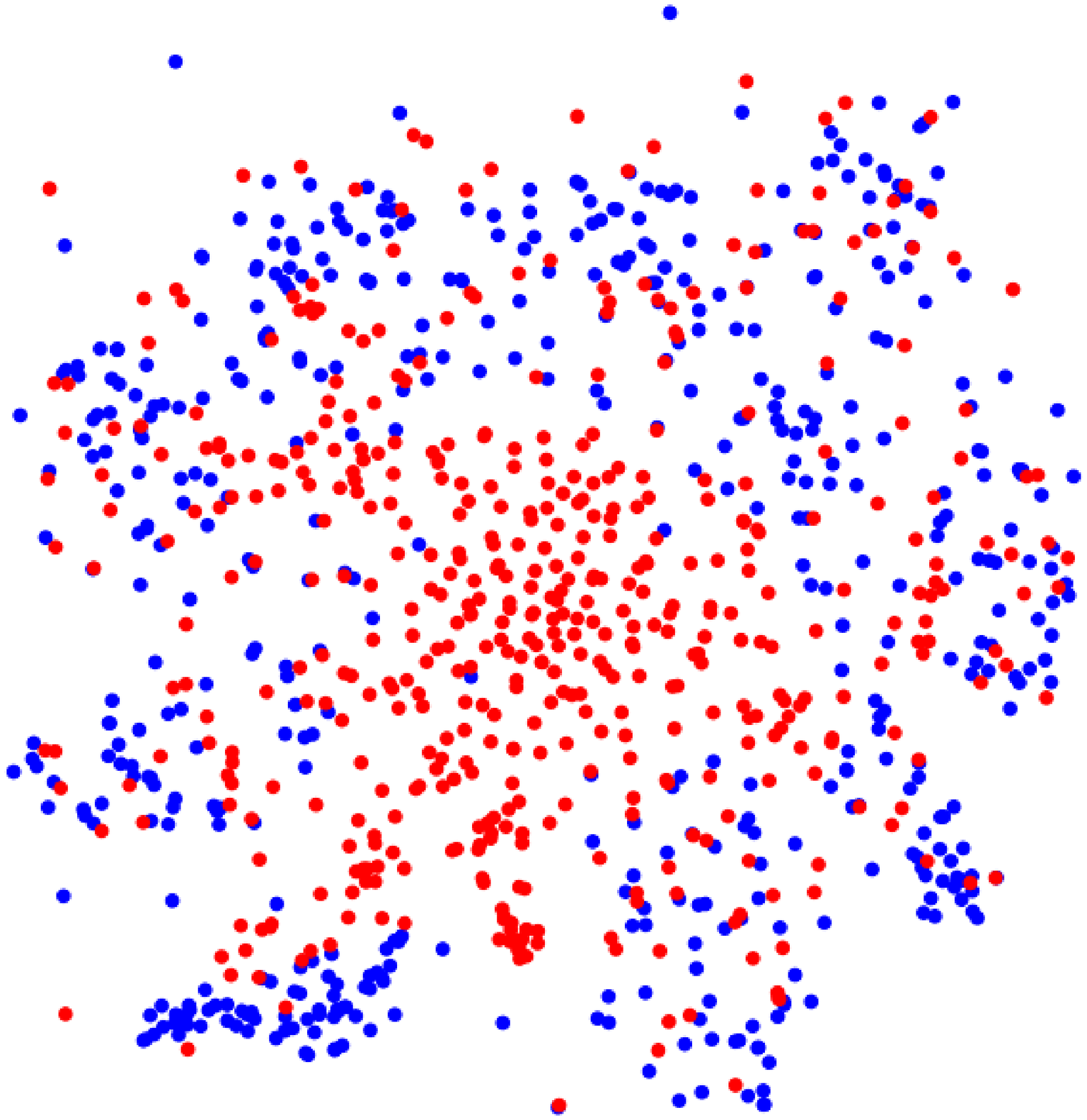}  \label{fig:mnist2m_ad}}
    \end{subfigure}
    \end{minipage}
    \begin{minipage}{0.5\hsize}
\subcaption{MNIST$\rightarrow$SVHN: last shared hidden layer}
  \centering
   \begin{subfigure}[Non-adapted]{
    \centering
   \includegraphics[width=0.45\hsize]{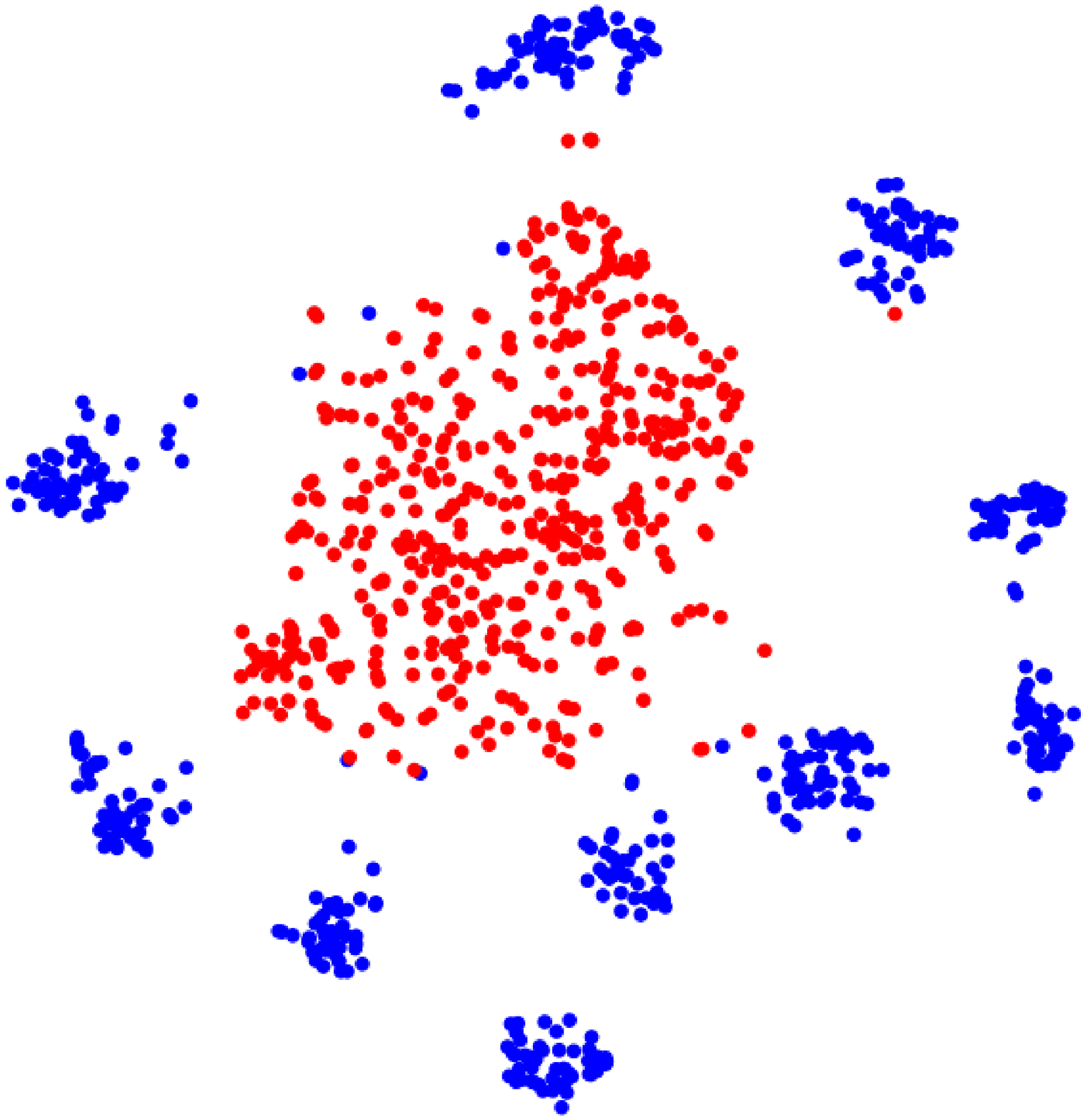}\label{fig:mnist2svhn_noad}}
    \end{subfigure}
      \begin{subfigure}[Adapted]{
    \centering
   \includegraphics[width=0.45\hsize]{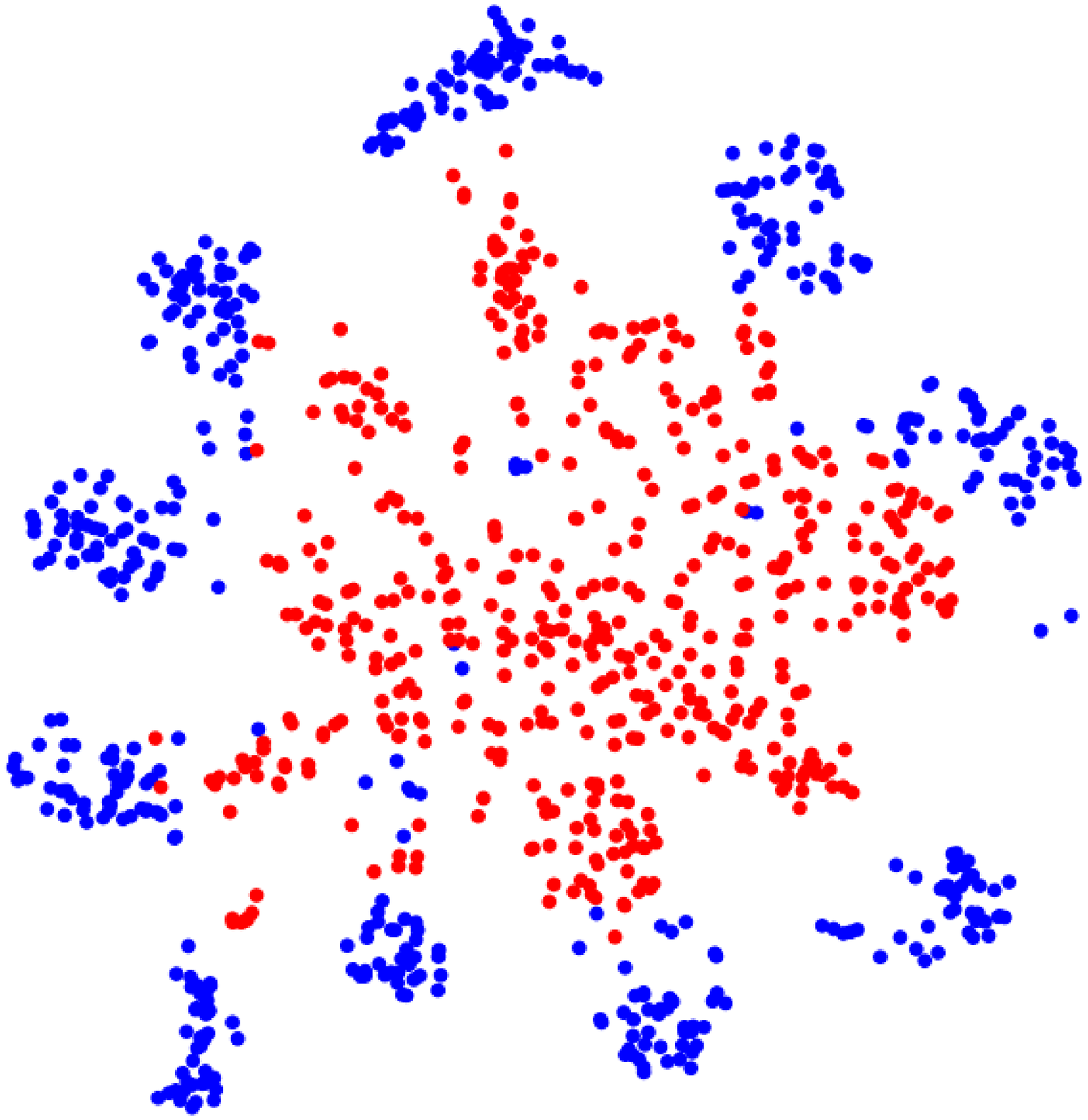}\label{fig:mnist2svhn_ad}}
    \end{subfigure}
    \end{minipage}
  \caption{We confirm the effect our method by visualization of the learned representations by using $t$-distributed stochastic neighbor embedding (t-SNE) \cite{maaten2008visualizing}. Red points are target samples and blue points are source samples. The samples are all from testing samples. {\bf (a), (c)} The case where we only use source samples for training. {\bf (b), (d)} The case of adaptation by our method. In both scenarios, MNIST$\rightarrow$SVHN and MNIST$\rightarrow$MNIST-M, we can see that the target samples are more dispersed through adaptation.}
  \label{fig:embed_mnist}
\end{figure*}
\vspace{-2mm}
\section{Experiment and Evaluation}
We perform extensive evaluations of our method on image datasets and a sentiment analysis dataset. We evaluate the accuracy of target-specific networks in all experiments. 

\textbf{Visual Domain Adaptation}
For visual domain adaptation, we perform our evaluation on the digits datasets and traffic signs datasets. Digits datasets include MNIST \cite{lecun1998gradient}, MNIST-M \cite{ganin2014unsupervised}, Street View House Numbers (SVHN) \cite{netzer2011reading}, and Synthetic Digits (SYN DIGITS) \cite{ganin2014unsupervised}. We further evaluate our method on traffic sign datasets including Synthetic Traffic Signs (SYN SIGNS) \cite{moiseev2013evaluation} and German Traffic Signs Recognition Benchmark \cite{stallkamp2011german} (GTSRB). In total, five adaptation scenarios are evaluated in this experiment. As the datasets used for evaluation are varied in previous works, we extensively evaluate our method on the five scenarios.

We do not evaluate our method on Office \cite{saenko2010adapting}, which is the most commonly used dataset for visual domain adaptation. As pointed out by \cite{bousmalis2016domain}, some labels in that dataset are noisy and some images contain other classes' objects.
Furthermore, many previous studies have evaluated the fine-tuning of pretrained networks using ImageNet. This protocol assumes the existence of another source domain. In our work, we want to evaluate the situation where we have access to only one source domain and one target domain.

\textbf{Adaptation in Amazon Reviews}
To investigate the behavior on language datasets, we also evaluated our method on the Amazon Reviews dataset \cite{blitzer2006domain} with the same preprocessing as used by \cite{coda,ganin2016domain}. The dataset contains reviews on four types of products: books, DVDs, electronics, and kitchen appliances.
We evaluate our method on 12 domain adaptation scenarios.
The results are shown in Table \ref{table:exp_scratch}.

\textbf{Baseline Methods}
We compare our method with five methods for unsupervised domain adaptation including state-of-the art methods in visual domain adaptation; Maximum Mean Discrepancy (MMD) \cite{long2015learning}, Domain Adversarial Neural Network (DANN) \cite{ganin2014unsupervised}, Deep Reconstruction Classification Network (DRCN) \cite{ghifary2016deep}, Domain Separation Networks (DSN) \cite{bousmalis2016domain}, and $k$-Nearest Neighbor based adaptation ($k$NN-Ad) \cite{sener2016learning}. We cite the results of MMD from \cite{bousmalis2016domain}. In addition, we compare our method with CNN trained only on source samples. We compare our method with Variational Fair AutoEncoder (VFAE) \cite{louizos2015variational} and DANN \cite{ganin2016domain} in the Amazon Reviews experiment. 

\subsection{Implementation Detail}
In experiments on image datasets, we employ the architecture of CNN used in \cite{ganin2014unsupervised}. For a fair comparison, we separate the network at the hidden layer from which \cite{ganin2014unsupervised} constructed discriminator networks. Therefore, when considering one classifier, for example, $F_1 \circ F$, the architecture is identical to previous work. We also follow \cite{ganin2014unsupervised} in the other protocols. We set the threshold value for the labeling method as 0.95 in MNIST$\rightarrow$SVHN. In other scenarios, we set it as 0.9. We use MomentumSGD for optimization and set the momentum as $0.9$, while the learning rate is determined on validation splits and uses either $[0.01, 0.05]$.
$\lambda$ is set 0.01 in all scenarios. In our Supplementary material, we provide details of the network architecture and hyper-parameters.

For experiments on the Amazon Reviews dataset, we use a similar architecture to that used in \cite{ganin2016domain}: with sigmoid activated, one dense hidden layer with 50 hidden units, and softmax output. We extend the architecture to our method similarly in the architecture of CNN. $\lambda$ is set as 0.001 based on the validation.
Since the input is sparse, we use Adagrad \cite{duchi2011adaptive} for optimization. We repeat this evaluation 10 times and report mean accuracy.
\begin{figure*}[h!]
\centering
   \begin{subfigure}[{\scriptsize MNIST$\rightarrow$MNIST-M}]{
   \centering
   \includegraphics[width=0.25\hsize]{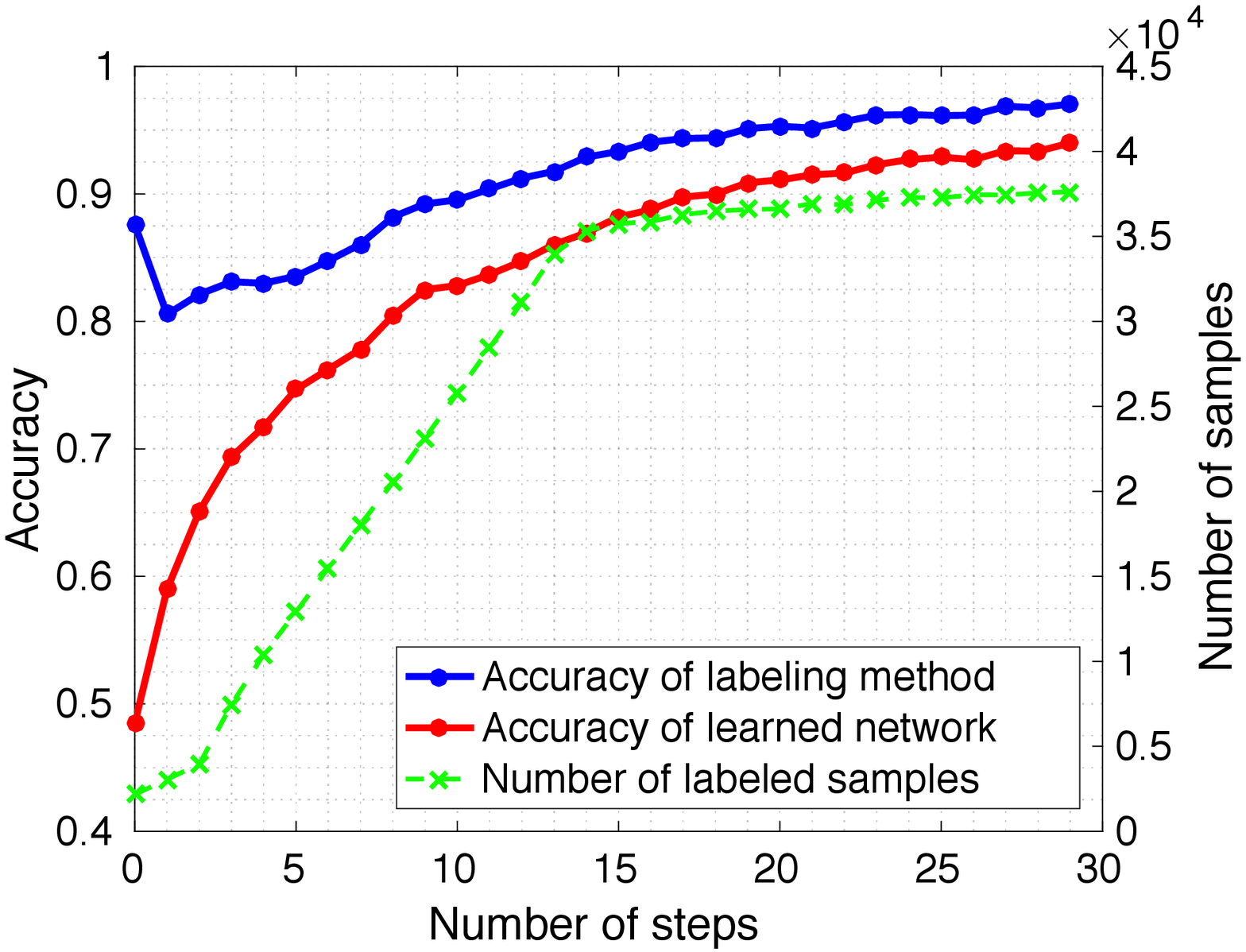} \label{fig:acc_mnistm}}
    \end{subfigure}
 \centering
   \begin{subfigure}[{\scriptsize SVHN$\rightarrow$MNIST}]{
   \centering
   \includegraphics[width=0.25\hsize]{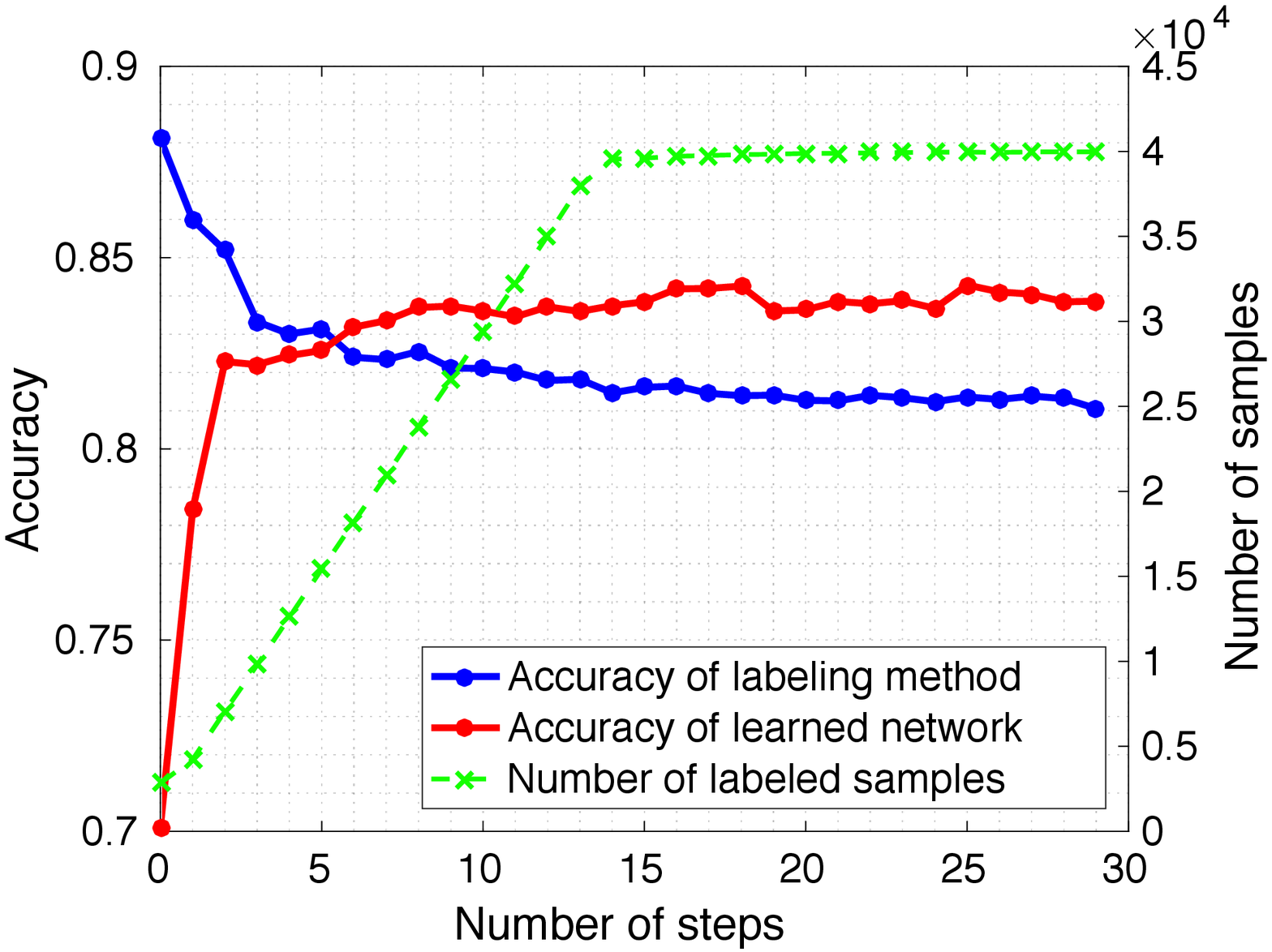} \label{fig:acc_mnist}}
    \end{subfigure}
\begin{subfigure}[{\scriptsize MNIST$\rightarrow$SVHN}]{
 \centering
   \includegraphics[width=0.25\hsize]{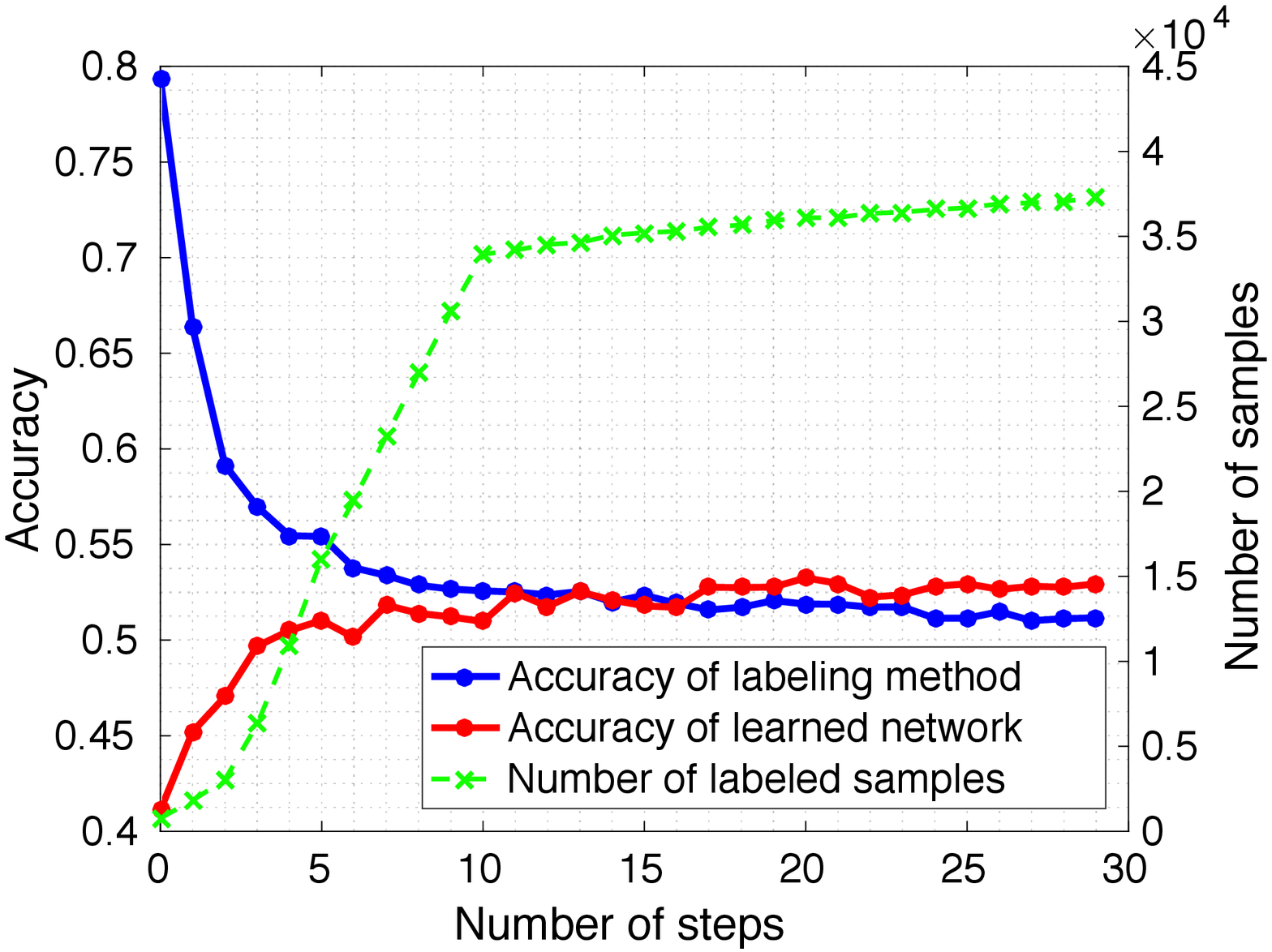}  \label{fig:acc_svhn}}
\end{subfigure}
\begin{subfigure}[{\scriptsize SYNDIGITS$\rightarrow$SVHN}]{
 \centering
   \includegraphics[width=0.24\hsize]{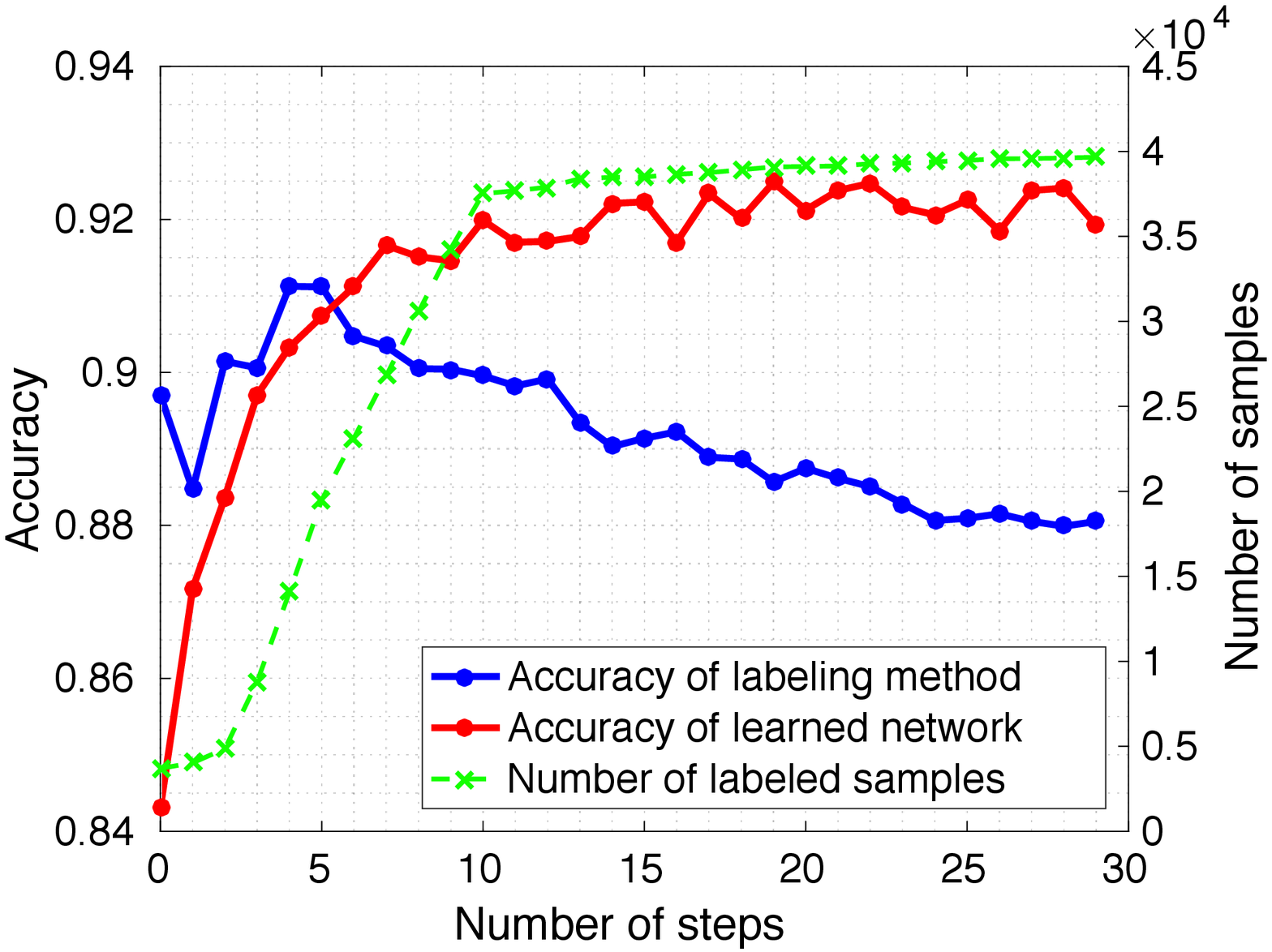}  \label{fig:acc_syn2svhn}}
\end{subfigure}
\begin{subfigure}[{\scriptsize SYNSIGNS$\rightarrow$GTSRB}]{
 \centering
   \includegraphics[width=0.24\hsize]{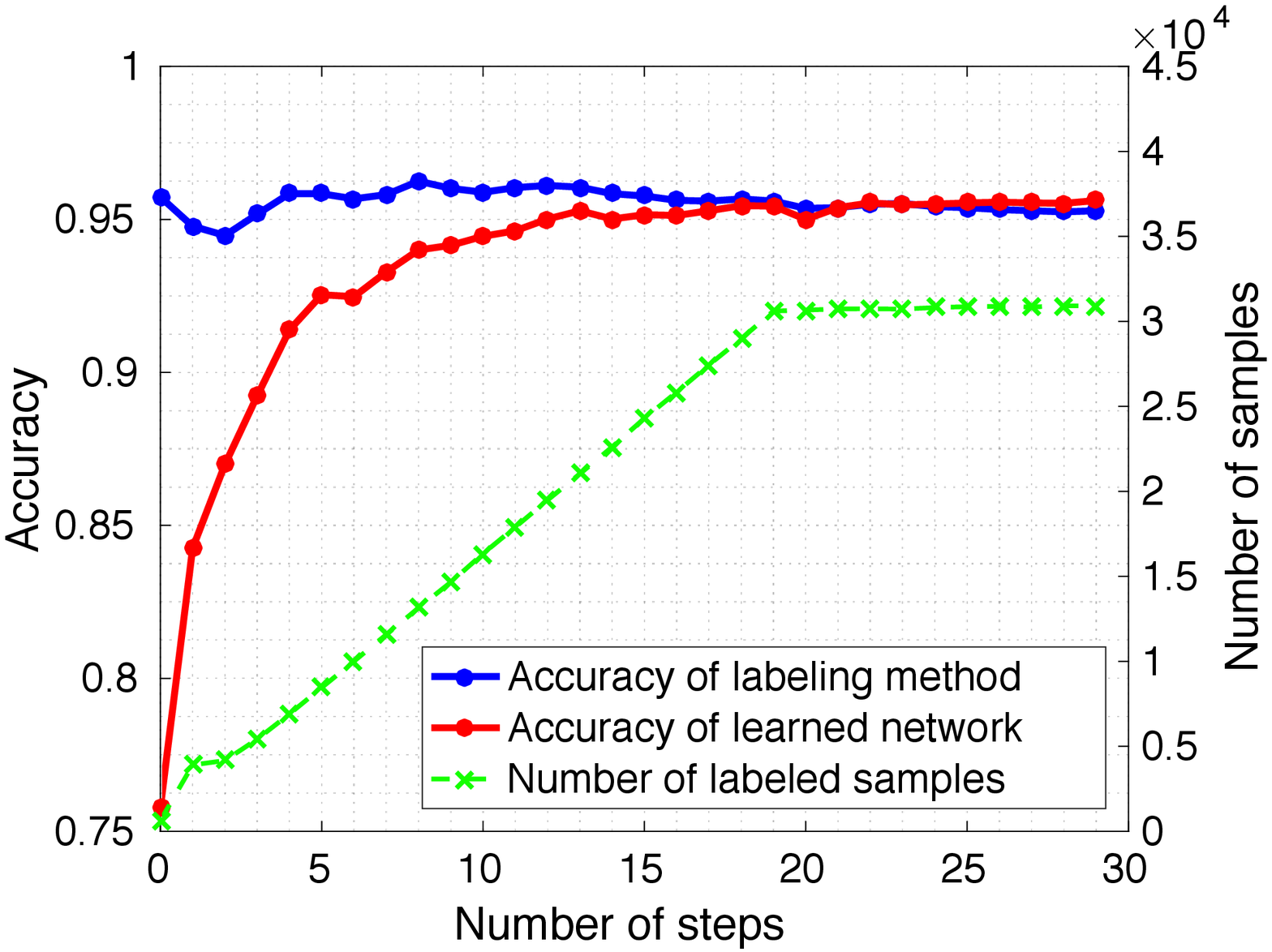}  \label{fig:acc_syn2gtr}}
\end{subfigure}
\centering
   \begin{subfigure}[{\scriptsize Comparision of accuracy of three network on SVHN$\rightarrow$MNIST}]{
   \centering
   \includegraphics[width=0.24\hsize]{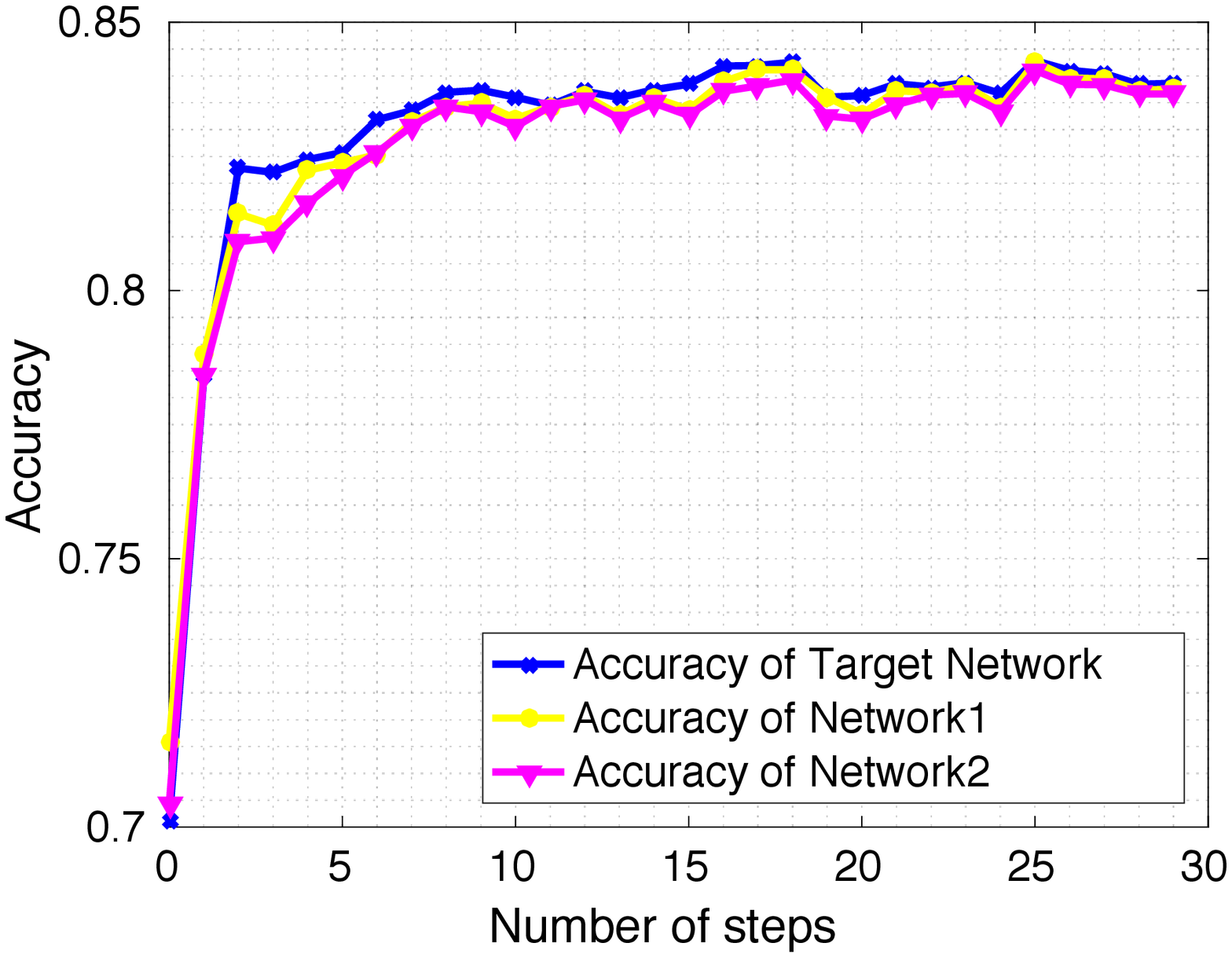} \label{fig:diff_net}}
   \end{subfigure}
   \centering
   \begin{subfigure}[{\scriptsize $\mathcal{A}$-distance in MNIST$\rightarrow$MNIST-M}]{
       \centering
   \includegraphics[width=0.22\hsize]{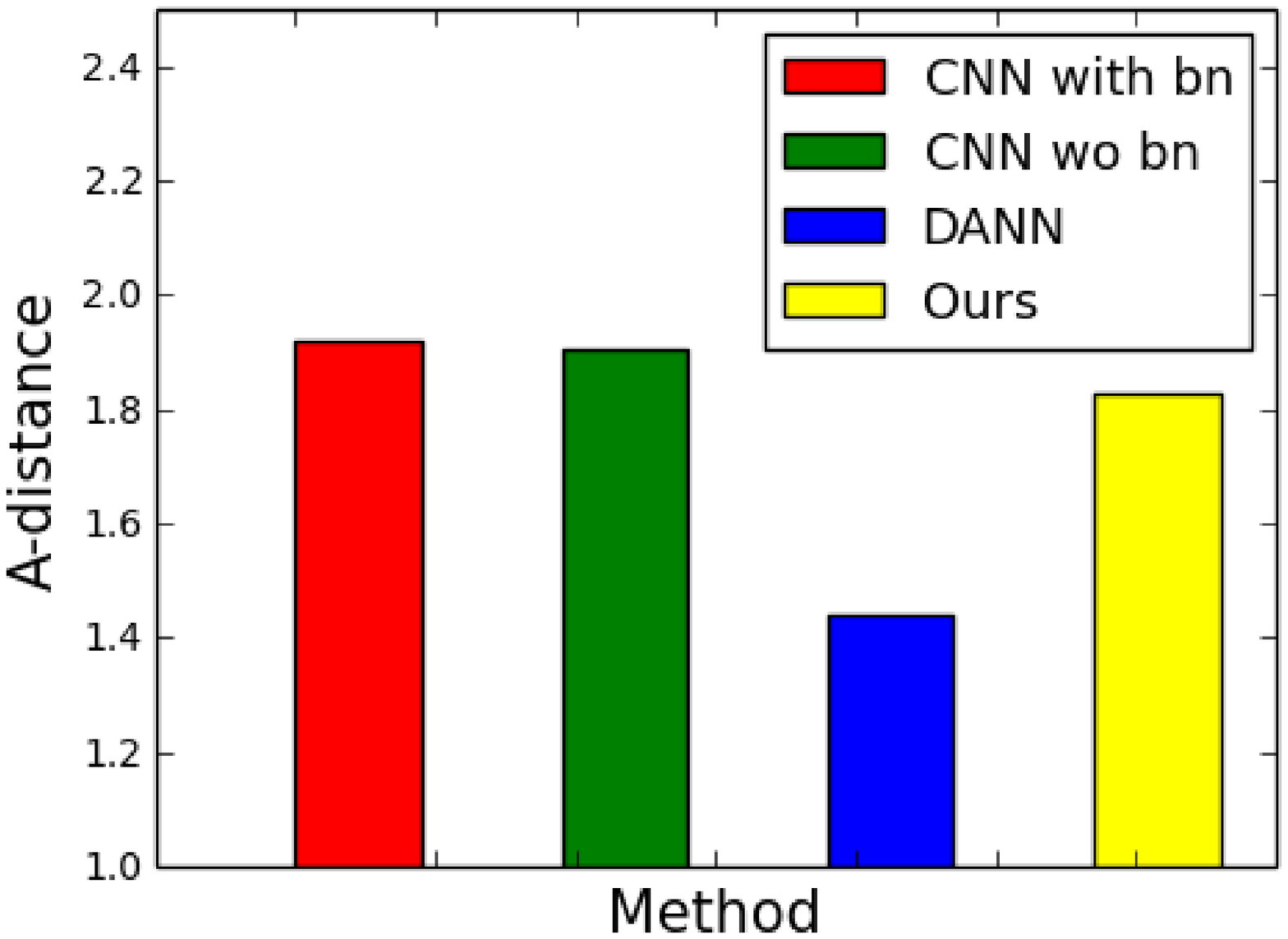} \label{fig:a_distance}}
    \end{subfigure}
    \caption{{\bf (a)} $\sim$ {\bf (e)}: Comparison of the actual accuracy of pseudo-labels and learned network accuracy during training. The blue curve is the pseudo-label accuracy and the red curve is the learned network accuracy. {Note that the labeling accuracy is computed using ({\it the number of correctly labeled samples})/({\it the number of labeled samples})}. The green curve is the number of labeled target samples in each step. {\bf (f)}: Comparison of the accuracy of three networks in our model. Three networks almost simultaneously improve accuracy. {\bf (g)}: Comparison of the $\mathcal{A}$-distance of different methods. Our model slightly reduced the divergence of the domain compared with source-only trained CNN.}
    \label{fig:acc_change}
    \end{figure*}

\subsection{Experimental Result}
In Tables \ref{table:exp_scratch} and \ref{table:amazon}, we show the main results of the experiments.
When training only on source samples, the effect of the BN is not clear as in Tables \ref{table:exp_scratch}.
However, in all image recognition experiments, the effect of BN in our method is clear; at the same time, the effect of our method is also clear when we do not use BN in the network architecture. The effect of the weight constraint is obvious in MNIST$\rightarrow$SVHN.

\textbf{MNIST$\rightarrow$MNIST-M}
First, we evaluate the adaptation scenario between the hand-written digits dataset MNIST and its transformed dataset MNIST-M. MNIST-M is composed by merging the clip of the background from BSDS500 datasets \cite{arbelaez2011contour}. A patch is randomly taken from the images in BSDS500, merged to MNIST digits. Even with this simple domain shift, the adaptation performance of CNN is much worse than the case where it was trained on target samples. From 59,001 target training samples, we randomly select 1,000 labeled target samples as a validation split and tuned hyper-parameters.

Our method outperforms the other existing method by about 7\%. Visualization of features in the last pooling layer is shown in Fig. \ref{fig:embed_mnist}\subref{fig:mnist2m_noad}\subref{fig:mnist2m_ad}. We can observe that the red target samples are more dispersed when adaptation is achieved. We show the comparison of the accuracy between the actual labeling accuracy on target samples during training and the test accuracy in Fig. \ref{fig:acc_change}. The test accuracy is very low at first, but as the steps increase, the accuracy becomes closer to that of the labeling accuracy. In this adaptation, we can clearly see that the actual labeling accuracy gradually improves with the accuracy of the network.

\textbf{SVHN$\leftrightarrow$MNIST}
We increase the gap between distributions in this experiment. We evaluate adaptation between SVHN \cite{netzer2011reading} and MNIST in a ten-class classification problem. SVHN and MNIST have distinct appearance, thus this adaptation is a challenging scenario especially in MNIST$\rightarrow$SVHN. SVHN is colored and some images contain multiple digits. Therefore, a classifier trained on SVHN is expected to perform well on MNIST, but the reverse is not true. MNIST does not include any samples containing multiple digits and most samples are centered in images, thus adaptation from MNIST to SVHN is rather difficult. In both settings, we use 1,000 labeled target samples to find the optimal hyperparameters.

We evaluate our method on both adaptation scenarios and achieved state-of-the-art performance on both datasets. In particular, for the adaptation MNIST$\rightarrow$SVHN, we outperformed other methods by more than 10\%. In Fig. \ref{fig:embed_mnist}\subref{fig:mnist2svhn_noad}\subref{fig:mnist2svhn_ad}, we visualize the representations in MNIST$\rightarrow$SVHN. Although the distributions seem to be separated between domains, the red SVHN samples become more discriminative using our method compared with non-adapted embedding. We also show the comparison between actual labeling method accuracy and testing accuracy in Fig. \ref{fig:acc_change}\subref{fig:acc_mnist}\subref{fig:acc_svhn}. In this figure, we can see that the labeling accuracy rapidly drops in the initial adaptation stage. On the other hand, testing accuracy continues to improve, and finally exceeds the labeling accuracy.
There are two questions about this interesting phenomenon. The first question is why does the labeling method continue to decrease despite the increase in the test accuracy? Target samples given pseudo-labels always include mistakenly labeled samples whereas those given no labels are ignored in our method. Therefore, the error will be reinforced in the target samples that are included in training set. The second question is why does the test accuracy continue to increase despite the lower labeling accuracy? The assumed reasons are that the network already acquires target discriminative representations in this phase and they can improve the accuracy using source samples and correctly labeled target samples.

In Fig. \ref{fig:diff_net}, we also show the comparison of accuracy of the three networks $F_1,F_2,F_t$ in SVHN$\rightarrow$MNIST. The accuracy of three networks is nearly the same in every step. The same thing is observed in other scenarios. From this result, we can state that the target-discriminative representations are shared in all three networks. 

\textbf{SYN DIGITS$\rightarrow$SVHN}
In this experiment, we aimed to address a common adaptation scenario from synthetic images to real images. The datasets of synthetic numbers \cite{ganin2014unsupervised} consist of 500,000 images generated from Windows fonts by varying the text, positioning, orientation, background and stroke colors, and the amount of blur. We use 479,400 source samples and 73,257 target samples for training, and 26,032 target samples for testing. We use 1,000 SVHN samples as a validation set.

Our method also outperforms other methods in this experiment.  In this experiment, the effect of BN is not clear compared with other scenarios. The domain gap is considered small in this scenario as the performance of the source-only classifier shows. In Fig. \ref{fig:acc_syn2svhn}, although the labeling accuracy is dropping, the accuracy of the learned network's prediction is improving as in MNIST$\leftrightarrow$SVHN.

\textbf{SYN SIGNS$\rightarrow$GTSRB}
This setting is similar to the previous setting, adaptation from synthetic images to real images, but we have a larger number of classes, namely 43 classes instead of 10. We use the SYN SIGNS dataset \cite{ganin2014unsupervised} for the source dataset and the GTSRB dataset \cite{stallkamp2011german} for the target dataset, which consist of real traffic sign images. We select randomly 31,367 samples for target training samples and evaluate accuracy on the rest of the samples. A total of 3,000 labeled target samples are used for validation.

In this scenario, our method outperforms other methods. This result shows that our method is effective for the adaptation from synthesized images to real images, which have diverse classes. In Fig. \ref{fig:acc_syn2gtr}, the same tendency as in MNIST$\leftrightarrow$SVHN is observed in this adaptation scenario.
\begin{table}
\begin{center}
\begin{tabular}{c|ccc}
\hline
\abovespace\belowspace
{Gradient stop branch}&{\scriptsize $F_t$}&{\scriptsize $F_1,F_2$}&{\scriptsize	None}\\\hline
 MNIST$\rightarrow$MNIST-M&56.4&95.4&94.0\\\hline
 MNIST$\rightarrow$SVHN&47.7&47.5&52.8\\\hline
 SYN SIGNS$\rightarrow$GTRB&96.5&93.1&96.2\\\hline
\end{tabular}
\caption{Results of Gradient stop experiment. When stopping gradients from $F_t$, we do not use backward gradients from $F_t$ to $F$, and $F$ learns only from $F_1,F_2$. When stopping gradients from $F_1,F_2$, we do not use backward gradients from $F_1,F_2$ to $F$, and $F$ learns from $F_t$. \textit{None} denotes our proposed method, we backward all gradients from all branches to $F$. In these three adaptation scenarios, our method shows stable performance.}
\label{table:stop_grad}
\end{center}
\end{table}

\textbf{Gradient Stop Experiment}
We evaluate the effect of the target-specific network in our method. We stop the gradient from upper layer networks $F_1,F_2$, and $F_t$ to examine the effect of $F_t$. Table \ref{table:stop_grad} shows three scenarios including the case where we stop the gradient from $F_1,F_2$, and $F_t$. In all scenarios, when we backward all gradients from $F_1,F_2,F_t$, we obtain clear performance improvements.

In the experiment MNIST$\rightarrow$MNIST-M, we can assume that only the backpropagation from $F_1,F_2$ cannot construct discriminative representations for target samples and confirm the effect of $F_t$.
For the adaptation MNIST$\rightarrow$SVHN, the best performance is realized when $F$ receives all gradients from upper networks. Backwarding all gradients will ensure both target-specific discriminative representations in difficult adaptations.
In SYN SIGNS$\rightarrow$GTSRB, backwarding only from $F_t$ produces the worst performance because these domains are similar and noisy pseudo-labeled target samples worsen the performance.

\textbf{$\mathcal{A}$-distance}
From the theoretical results in \cite{ben2010theory}, $\mathcal{A}$-distance is usually used as a measure of domain discrepancy. The way of estimating empirical $\mathcal{A}$-distance is simple, in which we train a classifier to classify a domain from each domains' feature. Then, the approximate distance is calculated as $\hat d_{\mathcal{A}} = 2(1-2\epsilon)$, where $\epsilon$ is the generalization error of the classifier. 
In Fig. \ref{fig:a_distance}, we show the $\mathcal{A}$-distance calculated from each CNN features. We used linear SVM to calculate the distance. From this graph, we can see that our method certainly reduces the $\mathcal{A}$-distance compared with the CNN trained on only source samples. In addition, when comparing DANN and our method, although DANN reduces $\mathcal{A}$-distance much more than our method, our method shows superior performance. This indicates that minimizing the domain discrepancy is not necessarily an appropriate way to achieve better performance.

\textbf{Amazon Reviews}
Reviews are encoded in 5,000 dimensional vectors of bag-of-words unigrams and bigrams with binary labels. Negative labels are attached to the samples if they are ranked with 1--3 stars. Positive labels are attached if they are ranked with 4 or 5 stars. We have 2,000 labeled source samples and 2,000 unlabeled target samples for training, and between 3,000 and 6,000 samples for testing. We use 200 of labeled target samples for validation.

From the results in Table \ref{table:amazon}, our method performs better than VFAE \cite{louizos2015variational} and DANN \cite{ganin2016domain} in nine settings out of twelve. Our method is effective in learning a shallow network on different domains. 
\begin{table}
\begin{center}
\begin{tabular}{c||c|c|c}
\toprule
\multirow{2}{*}{Source$\rightarrow$Target}&\multirow{2}{*}{VFAE}&\multirow{2}{*}{DANN}&\multirow{2}{*}{Our method}\\
&&&\\\midrule
 books$\rightarrow$dvd&79.9&78.4&{\bf 80.7}\\
 books$\rightarrow$electronics&79.2&73.3&{\bf 79.8}\\
 books$\rightarrow$kitchen&81.6&77.9&{\bf82.5}\\
 dvd$\rightarrow$books&{\bf75.5}&72.3&73.2\\
 dvd$\rightarrow$electronics&{\bf 78.6}&75.4&77.0\\
 dvd$\rightarrow$kitchen&82.2&78.3&{\bf 82.5}\\
 electronics$\rightarrow$books&72.7&71.1&{\bf73.2}\\
 electronics$\rightarrow$dvd&{\bf76.5}&73.8&72.9\\
 electronics$\rightarrow$kitchen&85.0&85.4&{\bf 86.9}\\
 kitchen$\rightarrow$books&72.0&70.9&{\bf 72.5}\\
 kitchen$\rightarrow$dvd&73.3&74.0&{\bf 74.9}\\
 kitchen$\rightarrow$electronics&83.8&84.3&{\bf 84.6}\\\bottomrule
\end{tabular}
\caption{Amazon Reviews experimental results. The accuracy (\%) of the proposed method is shown with the result of VFAE \cite{louizos2015variational} and DANN \cite{ganin2016domain}.}
\label{table:amazon}
\end{center}
\end{table}
\vspace{-2mm}
\section{Conclusion}
In this paper, we have proposed a novel asymmetric tri-training method for unsupervised domain adaptation, which is simply implemented. We aimed to learn discriminative representations by utilizing pseudo-labels assigned to unlabeled target samples. 
We utilized three classifiers, two networks assign pseudo-labels to unlabeled target samples and the remaining network learns from them.
We evaluated our method both on domain adaptation on a visual recognition task and a sentiment analysis task, outperforming other methods. In particular, our method outperformed the other methods by more than 10\% in the MNIST$\rightarrow$SVHN adaptation task.
\section{Acknowledgement}
This work was funded by ImPACT Program of Council for Science, Technology and Innovation (Cabinet Office, Government of Japan) and supported by CREST, JST.

\bibliography{example_paper}

\begin{thebibliography}{40}
\providecommand{\natexlab}[1]{#1}
\providecommand{\url}[1]{\texttt{#1}}
\expandafter\ifx\csname urlstyle\endcsname\relax
  \providecommand{\doi}[1]{doi: #1}\else
  \providecommand{\doi}{doi: \begingroup \urlstyle{rm}\Url}\fi

\bibitem[Antol et~al.(2015)Antol, Agrawal, Lu, Mitchell, Batra,
  Lawrence~Zitnick, and Parikh]{antol2015vqa}
Antol, Stanislaw, Agrawal, Aishwarya, Lu, Jiasen, Mitchell, Margaret, Batra,
  Dhruv, Lawrence~Zitnick, C., and Parikh, Devi.
\newblock Vqa: Visual question answering.
\newblock In \emph{ICCV}, 2015.

\bibitem[Arbelaez et~al.(2011)Arbelaez, Maire, Fowlkes, and
  Malik]{arbelaez2011contour}
Arbelaez, Pablo, Maire, Michael, Fowlkes, Charless, and Malik, Jitendra.
\newblock Contour detection and hierarchical image segmentation.
\newblock \emph{PAMI}, 33\penalty0 (5):\penalty0 898--916, 2011.

\bibitem[Balcan et~al.(2004)Balcan, Blum, and Yang]{balcan2004co}
Balcan, Maria-Florina, Blum, Avrim, and Yang, Ke.
\newblock Co-training and expansion: Towards bridging theory and practice.
\newblock In \emph{NIPS}, 2004.

\bibitem[Ben-David et~al.(2010)Ben-David, Blitzer, Crammer, Kulesza, Pereira,
  and Vaughan]{ben2010theory}
Ben-David, Shai, Blitzer, John, Crammer, Koby, Kulesza, Alex, Pereira,
  Fernando, and Vaughan, Jennifer~Wortman.
\newblock A theory of learning from different domains.
\newblock \emph{Machine learning}, 79\penalty0 (1-2):\penalty0 151--175, 2010.

\bibitem[Blitzer et~al.(2006)Blitzer, McDonald, and Pereira]{blitzer2006domain}
Blitzer, John, McDonald, Ryan, and Pereira, Fernando.
\newblock Domain adaptation with structural correspondence learning.
\newblock In \emph{EMNLP}, 2006.

\bibitem[Blum \& Mitchell(1998)Blum and Mitchell]{blum1998combining}
Blum, Avrim and Mitchell, Tom.
\newblock Combining labeled and unlabeled data with co-training.
\newblock In \emph{COLT}, 1998.

\bibitem[Bousmalis et~al.(2016)Bousmalis, Trigeorgis, Silberman, Krishnan, and
  Erhan]{bousmalis2016domain}
Bousmalis, Konstantinos, Trigeorgis, George, Silberman, Nathan, Krishnan,
  Dilip, and Erhan, Dumitru.
\newblock Domain separation networks.
\newblock In \emph{NIPS}, 2016.

\bibitem[Chen et~al.(2011)Chen, Weinberger, and Blitzer]{coda}
Chen, Minmin, Weinberger, Kilian~Q, and Blitzer, John.
\newblock Co-training for domain adaptation.
\newblock In \emph{NIPS}, 2011.

\bibitem[Dasgupta et~al.(2001)Dasgupta, Littman, and
  McAllester]{dasgupta2001pac}
Dasgupta, Sanjoy, Littman, Michael~L, and McAllester, David.
\newblock Pac generalization bounds for co-training.
\newblock In \emph{NIPS}, 2001.

\bibitem[Deng et~al.(2009)Deng, Dong, Socher, Li, Li, and
  Fei-Fei]{deng2009imagenet}
Deng, Jia, Dong, Wei, Socher, Richard, Li, Li-Jia, Li, Kai, and Fei-Fei, Li.
\newblock Imagenet: A large-scale hierarchical image database.
\newblock In \emph{CVPR}, 2009.

\bibitem[Duchi et~al.(2011)Duchi, Hazan, and Singer]{duchi2011adaptive}
Duchi, John, Hazan, Elad, and Singer, Yoram.
\newblock Adaptive subgradient methods for online learning and stochastic
  optimization.
\newblock \emph{JMLR}, 12\penalty0 (7):\penalty0 2121--2159, 2011.

\bibitem[Ganin \& Lempitsky(2014)Ganin and Lempitsky]{ganin2014unsupervised}
Ganin, Yaroslav and Lempitsky, Victor.
\newblock Unsupervised domain adaptation by backpropagation.
\newblock In \emph{ICML}, 2014.

\bibitem[Ganin et~al.(2016)Ganin, Ustinova, Ajakan, Germain, Larochelle,
  Laviolette, Marchand, and Lempitsky]{ganin2016domain}
Ganin, Yaroslav, Ustinova, Evgeniya, Ajakan, Hana, Germain, Pascal, Larochelle,
  Hugo, Laviolette, Fran{\c{c}}ois, Marchand, Mario, and Lempitsky, Victor.
\newblock Domain-adversarial training of neural networks.
\newblock \emph{JMLR}, 17\penalty0 (59):\penalty0 1--35, 2016.

\bibitem[Ghifary et~al.(2016)Ghifary, Kleijn, Zhang, Balduzzi, and
  Li]{ghifary2016deep}
Ghifary, Muhammad, Kleijn, W~Bastiaan, Zhang, Mengjie, Balduzzi, David, and Li,
  Wen.
\newblock Deep reconstruction-classification networks for unsupervised domain
  adaptation.
\newblock In \emph{ECCV}, 2016.

\bibitem[Girshick et~al.(2014)Girshick, Donahue, Darrell, and
  Malik]{girshick2014rich}
Girshick, Ross, Donahue, Jeff, Darrell, Trevor, and Malik, Jitendra.
\newblock Rich feature hierarchies for accurate object detection and semantic
  segmentation.
\newblock In \emph{CVPR}, 2014.

\bibitem[Gretton et~al.(2012)Gretton, Borgwardt, Rasch, Sch{\"o}lkopf, and
  Smola]{gretton2012kernel}
Gretton, Arthur, Borgwardt, Karsten~M, Rasch, Malte~J, Sch{\"o}lkopf, Bernhard,
  and Smola, Alexander.
\newblock A kernel two-sample test.
\newblock \emph{JMLR}, 13\penalty0 (3):\penalty0 723--773, 2012.

\bibitem[Ioffe \& Szegedy(2015)Ioffe and Szegedy]{ioffe2015batch}
Ioffe, Sergey and Szegedy, Christian.
\newblock Batch normalization: Accelerating deep network training by reducing
  internal covariate shift.
\newblock \emph{arXiv:1502.03167}, 2015.

\bibitem[Khamis \& Lampert(2014)Khamis and Lampert]{khamis2014coconut}
Khamis, Sameh and Lampert, Christoph~H.
\newblock Coconut: Co-classification with output space regularization.
\newblock In \emph{BMVC}, 2014.

\bibitem[Krizhevsky et~al.(2012)Krizhevsky, Sutskever, and
  Hinton]{krizhevsky2012imagenet}
Krizhevsky, Alex, Sutskever, Ilya, and Hinton, Geoffrey~E.
\newblock Imagenet classification with deep convolutional neural networks.
\newblock In \emph{NIPS}, 2012.

\bibitem[LeCun et~al.(1998)LeCun, Bottou, Bengio, and
  Haffner]{lecun1998gradient}
LeCun, Yann, Bottou, L{\'e}on, Bengio, Yoshua, and Haffner, Patrick.
\newblock Gradient-based learning applied to document recognition.
\newblock \emph{Proceedings of the IEEE}, 86\penalty0 (11):\penalty0
  2278--2324, 1998.

\bibitem[Lee(2013)]{lee2013pseudo}
Lee, Dong-Hyun.
\newblock Pseudo-label: The simple and efficient semi-supervised learning
  method for deep neural networks.
\newblock In \emph{ICML workshop on Challenges in Representation Learning},
  2013.

\bibitem[Levin et~al.(2003)Levin, Viola, and Freund]{levin2003unsupervised}
Levin, Anat, Viola, Paul~A, and Freund, Yoav.
\newblock Unsupervised improvement of visual detectors using co-training.
\newblock In \emph{ICCV}, 2003.

\bibitem[Li et~al.(2016)Li, Wang, Shi, Liu, and Hou]{li2016revisiting}
Li, Yanghao, Wang, Naiyan, Shi, Jianping, Liu, Jiaying, and Hou, Xiaodi.
\newblock Revisiting batch normalization for practical domain adaptation.
\newblock \emph{arXiv:1603.04779}, 2016.

\bibitem[Long et~al.(2015{\natexlab{a}})Long, Shelhamer, and
  Darrell]{long2015fully}
Long, Jonathan, Shelhamer, Evan, and Darrell, Trevor.
\newblock Fully convolutional networks for semantic segmentation.
\newblock In \emph{CVPR}, 2015{\natexlab{a}}.

\bibitem[Long et~al.(2015{\natexlab{b}})Long, Cao, Wang, and
  Jordan]{long2015learning}
Long, Mingsheng, Cao, Yue, Wang, Jianmin, and Jordan, Michael~I.
\newblock Learning transferable features with deep adaptation networks.
\newblock In \emph{ICML}, 2015{\natexlab{b}}.

\bibitem[Long et~al.(2016)Long, Zhu, Wang, and Jordan]{long2016unsupervised}
Long, Mingsheng, Zhu, Han, Wang, Jianmin, and Jordan, Michael~I.
\newblock Unsupervised domain adaptation with residual transfer networks.
\newblock In \emph{NIPS}, 2016.

\bibitem[Louizos et~al.(2015)Louizos, Swersky, Li, Welling, and
  Zemel]{louizos2015variational}
Louizos, Christos, Swersky, Kevin, Li, Yujia, Welling, Max, and Zemel, Richard.
\newblock The variational fair autoencoder.
\newblock \emph{arXiv:1511.00830}, 2015.

\bibitem[Maaten \& Hinton(2008)Maaten and Hinton]{maaten2008visualizing}
Maaten, Laurens van~der and Hinton, Geoffrey.
\newblock Visualizing data using t-sne.
\newblock \emph{JMLR}, 9\penalty0 (11):\penalty0 2579--2605, 2008.

\bibitem[Moiseev et~al.(2013)Moiseev, Konev, Chigorin, and
  Konushin]{moiseev2013evaluation}
Moiseev, Boris, Konev, Artem, Chigorin, Alexander, and Konushin, Anton.
\newblock Evaluation of traffic sign recognition methods trained on
  synthetically generated data.
\newblock In \emph{ACIVS}, 2013.

\bibitem[Netzer et~al.(2011)Netzer, Wang, Coates, Bissacco, Wu, and
  Ng]{netzer2011reading}
Netzer, Yuval, Wang, Tao, Coates, Adam, Bissacco, Alessandro, Wu, Bo, and Ng,
  Andrew~Y.
\newblock Reading digits in natural images with unsupervised feature learning.
\newblock In \emph{NIPS workshop on deep learning and unsupervised feature
  learning}, 2011.

\bibitem[Rohrbach et~al.(2013)Rohrbach, Ebert, and
  Schiele]{rohrbach2013transfer}
Rohrbach, Marcus, Ebert, Sandra, and Schiele, Bernt.
\newblock Transfer learning in a transductive setting.
\newblock In \emph{NIPS}, 2013.

\bibitem[Saenko et~al.(2010)Saenko, Kulis, Fritz, and
  Darrell]{saenko2010adapting}
Saenko, Kate, Kulis, Brian, Fritz, Mario, and Darrell, Trevor.
\newblock Adapting visual category models to new domains.
\newblock In \emph{ECCV}, 2010.

\bibitem[Sener et~al.(2016)Sener, Song, Saxena, and
  Savarese]{sener2016learning}
Sener, Ozan, Song, Hyun~Oh, Saxena, Ashutosh, and Savarese, Silvio.
\newblock Learning transferrable representations for unsupervised domain
  adaptation.
\newblock In \emph{NIPS}, 2016.

\bibitem[Stallkamp et~al.(2011)Stallkamp, Schlipsing, Salmen, and
  Igel]{stallkamp2011german}
Stallkamp, Johannes, Schlipsing, Marc, Salmen, Jan, and Igel, Christian.
\newblock The german traffic sign recognition benchmark: a multi-class
  classification competition.
\newblock In \emph{IJCNN}, 2011.

\bibitem[Sun et~al.(2016)Sun, Feng, and Saenko]{sun2015return}
Sun, Baochen, Feng, Jiashi, and Saenko, Kate.
\newblock Return of frustratingly easy domain adaptation.
\newblock In \emph{AAAI}, 2016.

\bibitem[Tanha et~al.(2011)Tanha, van Someren, and
  Afsarmanesh]{tanha2011ensemble}
Tanha, Jafar, van Someren, Maarten, and Afsarmanesh, Hamideh.
\newblock Ensemble based co-training.
\newblock In \emph{BNAIC}, 2011.

\bibitem[Vinyals et~al.(2015)Vinyals, Toshev, Bengio, and
  Erhan]{vinyals2015show}
Vinyals, Oriol, Toshev, Alexander, Bengio, Samy, and Erhan, Dumitru.
\newblock Show and tell: A neural image caption generator.
\newblock In \emph{CVPR}, 2015.

\bibitem[Wan(2009)]{wan2009co}
Wan, Xiaojun.
\newblock Co-training for cross-lingual sentiment classification.
\newblock In \emph{ACL}, 2009.

\bibitem[Zhou \& Li(2005)Zhou and Li]{zhou2005tri}
Zhou, Zhi-Hua and Li, Ming.
\newblock Tri-training: Exploiting unlabeled data using three classifiers.
\newblock \emph{TKDE}, 17\penalty0 (11):\penalty0 1529--1541, 2005.

\bibitem[Zhu(2005)]{zhu2005semi}
Zhu, Xiaojin.
\newblock Semi-supervised learning literature survey.
\newblock Technical report, University of Wisconsin-Madison, 2005.

\end{thebibliography}
\bibliographystyle{icml2017}
\newpage
\section*{Proof of Theorem}
We introduce the derivation of theorem of the main paper.
The ideal joint hypothesis is defined as $h^{*}= \argmin_{h\in H}\bigl(R_{\mathcal{S}}(h^{*})+R_{\mathcal{T}}(h^{*})\bigr)$, and its corresponding error is $C=R_{\mathcal{S}}(h^{*})+R_{\mathcal{T}}(h^{*})$, where $R$ denotes the expected error on each hypothesis.

We consider the pseudo-labeled target samples set $T_{l}=\bigl\{(x_i,\hat{y}_i)\bigr\}^{m_t}_{i=1}$ given false labels at the ratio of $\rho$. The minimum shared error on $\mathcal{S},\mathcal{T}_{l}$ is denoted as $C{'}$. Then, the following inequality holds:

\begin{math}
\forall h  \in H, R_{\mathcal{T}}(h) \leq R_{\mathcal{S}}(h)  +\frac{1}{2}{d_{\mathcal{H} \Delta \mathcal{H}}(\mathcal{S}_{{\bf X}},\mathcal{T}_{{\bf X}})}+C\\\\
						   \ \ \ \ \ \ \ \ \ \ \ \ \ \ \ \ \ \ \ \ \ \ \ \ \ \ \leq	R_{\mathcal{S}}(h)  +\frac{1}{2}{d_{\mathcal{H} \Delta \mathcal{H}}(\mathcal{S}_{{\bf X}},\mathcal{T}_{{\bf X}})}+C{'}+\rho
\end{math}

\begin{proof}
  The probabiliy of false labels in the pseudo-labeled set $T_{l}$ is $\rho$.
  When we consider 0-1 loss function for $l$, the difference between the error based on the true labeled set and pseudo-labeled set is
  \begin{eqnarray}
    |l(h(x_i),y_i) - l(h(x_i),\hat y_i)|=\begin{cases}
    \ \ 1 \ \ & y_i\neq \hat y_i\\\nonumber
    \ \ 0 \ \ & y_i = \hat y_i
    \end{cases}
  \end{eqnarray}\\
      Then, the difference in the expected error is,
  \begin{eqnarray}
    {\mathbf{E}}[|l(h(x_i),y_i) - l(h(x_i),\hat y_i)|]
    \leq |R_{\mathcal{T}_l}(h) -R_{\mathcal{T}}(h)| \leq \rho \nonumber
  \end{eqnarray}
  From the characteritic of the loss function, the triangle inequality will hold, then
  \begin{eqnarray}
    \nonumber
    R_{\mathcal{S}}(h) + R_{\mathcal{T}}(h)&=&R_{\mathcal{S}}(h) + R_{\mathcal{T}}(h) - R_{\mathcal{T}_{l}}(h) + R_{\mathcal{T}_{l}}(h) \\\nonumber
    &\leq& R_{\mathcal{S}}(h) + R_{\mathcal{T}_l}(h) +|R_{\mathcal{T}_l}(h) -R_{\mathcal{T}}(h)|\\\nonumber
    &\leq& R_{\mathcal{S}}(h) + R_{\mathcal{T}_{l}}(h) + \rho \nonumber
  \end{eqnarray}
  From this result, the main inequality holds.
 \end{proof}

\section*{CNN Architectures and training detail}
\begin{figure*}[t]
  \begin{center}
   \includegraphics[width=\hsize]{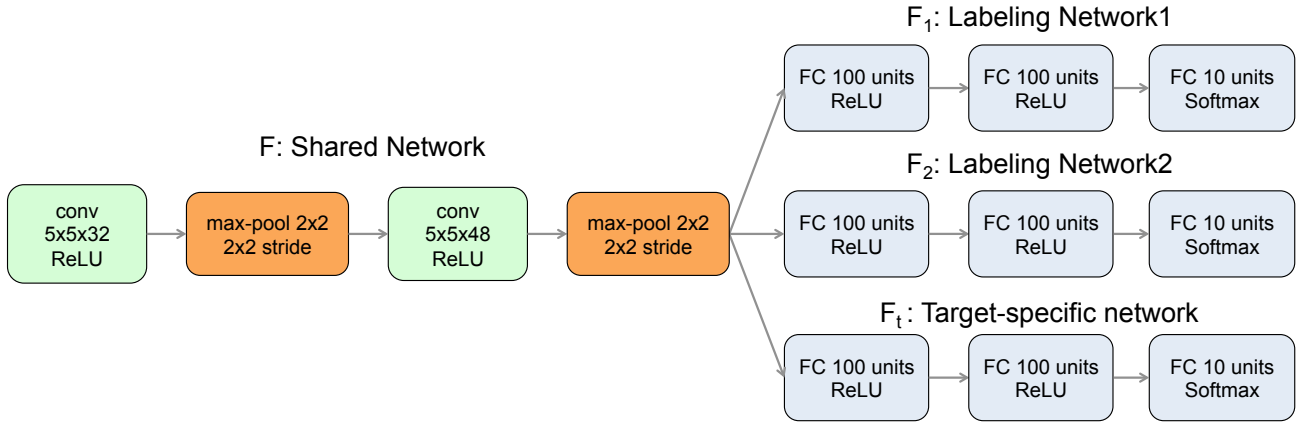}
  \end{center}
\caption{The architecture used for MNIST$\rightarrow$MNIST-M. We added BN layer in the last convolution layer and FC layers in $F_{1},F_{2}$. We also used dropout in our experiment.}
    \label{fig:mnist_arc}
\end{figure*}

\begin{figure*}[t]
  \begin{center}
   \includegraphics[width=\hsize]{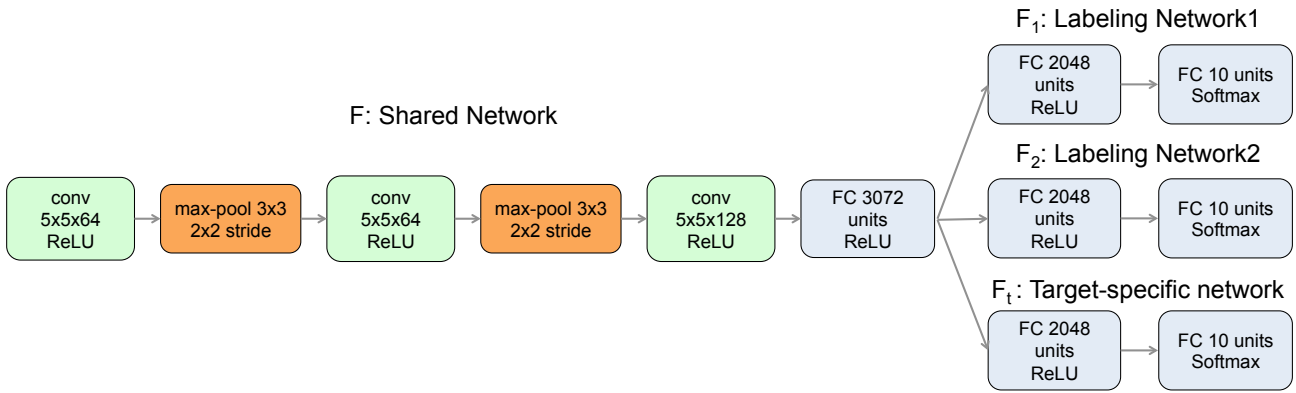}
  \end{center}
\caption{The architecture used for training SVHN. In MNIST$\rightarrow$SVHN, we added a BN layer in the last FC layer in $F$. In SVHN$\rightarrow$MNIST, SYN Digits$\leftrightarrow$SVHN, we added BN layer in the last convolution layer in $F$ and FC layers in $F_{1}$,$F_{2}$ and also used dropout.}
    \label{fig:svhn_arc}
\end{figure*}

\begin{figure*}[t]
  \begin{center}
   \includegraphics[width=\hsize]{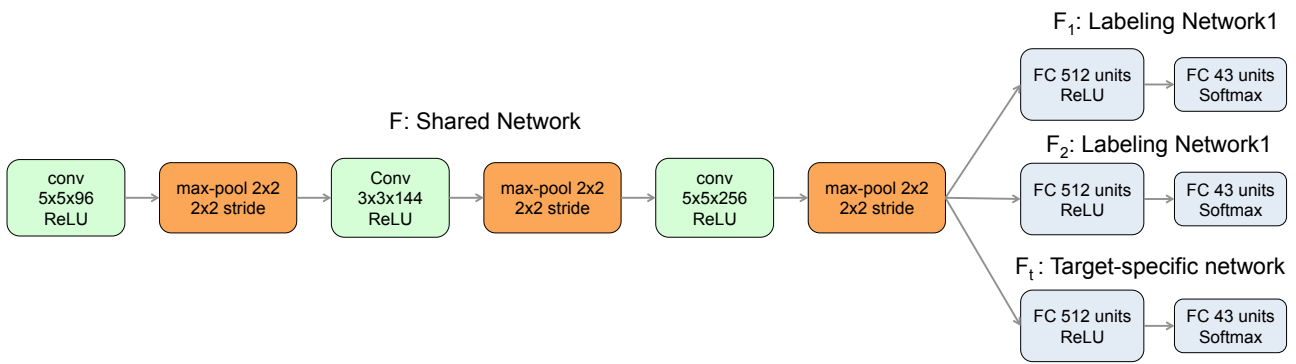}
  \end{center}
\caption{The architecture used in the adaptation Synthetic Signs$\rightarrow$GTSRB. We added a BN layer after the last convolution layer in $F$ and also used dropout.}
    \label{fig:gtsr_arc}
\end{figure*}

Four types of architectures are used for our method, which is based on \cite{ganin2014unsupervised}. The network topology is shown in Figs \ref{fig:mnist_arc}, \ref{fig:svhn_arc} and \ref{fig:gtsr_arc}. The other hyperparameters are decided on the validation splits. The learning rate is set to 0.05 in SVHN$\leftrightarrow$MNIST. In the other scenarios, it is set to 0.01. The batchsize for training $F_t,F$ is set as 128, the batchsize for training $F_1,F_2,F$ is set as 64 in all scenarios.

In MNIST$\rightarrow$MNIST-M, the dropout rate used in the experiment is 0.2 for training $F_t$, 0.5 for training $F_1,F_2$.
In MNIST$\rightarrow$SVHN, we did not use dropout. We decreased learning rate to 0.001 after step 10.
In SVHN$\rightarrow$MNIST, the dropout rate used in the experiment is 0.5.
In SYNDIGITS$\rightarrow$SVHN, the dropout rate used in the experiment is 0.5.
In SYNSIGNS$\rightarrow$GTSRB, the dropout rate used in the experiment is 0.5.
\section*{Supplementary experiments on MNIST$\rightarrow$MNIST-M}
\begin{figure}[t]
  \begin{center}
   \includegraphics[width=\hsize]{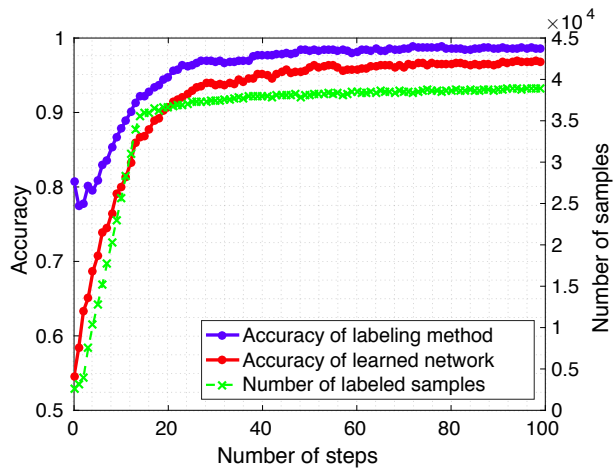}
  \end{center}
\caption{The behavior of our model when increasing the number of steps up to 100. Our model achieves accuracy of about 97\%.}
    \label{fig:100epoch}
\end{figure}

We observe the behavior of our model when increasing the number of steps up to one hundred. We show the result in Fig. \ref{fig:100epoch}. Our model's accuracy gets about 97\%. In our main experiments, we set the number of steps thirty, but from this experiment, further improvements can be expected.

\end{document}